\documentclass[conference]{IEEEtran}
\IEEEoverridecommandlockouts
\usepackage{cite}
\usepackage{amsmath,amssymb,amsfonts}
\usepackage{algorithmic}
\usepackage{graphicx}
\usepackage{textcomp}
\usepackage{xcolor}
\usepackage{color}
\usepackage[utf8]{inputenc}
\usepackage[T1]{fontenc}
\usepackage{amsmath,amssymb}
\usepackage{graphicx}
\usepackage{hyperref} 
\usepackage{cite}
\usepackage{booktabs}
\usepackage{algorithmic}
\usepackage{algorithm}
\usepackage{microtype}
\usepackage{xcolor}
\usepackage{lipsum}
\usepackage{placeins}
\usepackage{dblfloatfix}
\usepackage{tikz}
\usepackage{multirow}
\usepackage{placeins}
\usepackage{tcolorbox}
\usepackage{microtype}
\usepackage{tabularx}
\tcbuselibrary{skins, breakable}
\usetikzlibrary{positioning,calc,fit,shapes.callouts}
\usepackage{svg}
\usepackage{listings}
\usepackage{array}
\usepackage[table]{xcolor}
\newcommand{\subsecref}[2]{Section~\ref{#1}-\ref{#2}}
\newcommand{\rowrule}{%
  \arrayrulecolor{gray!25}%
  \specialrule{0.35pt}{2pt}{2pt}%
  \arrayrulecolor{black}%
}
\include{bib_ieee_abbr.def}
\def\BibTeX{{\rm B\kern-.05em{\sc i\kern-.025em b}\kern-.08em
    T\kern-.1667em\lower.7ex\hbox{E}\kern-.125emX}}
\begin{document}

\title{Speed in the Blind Spot: An Interpretability Analysis of Dynamic Perception in VLMs for Autonomous Driving}

\author{Katharina Winter$^{1}$, Stefan Englmeier$^{1}$, Fabian B. Flohr$^{1}$
\thanks{$^{1}$Munich University of Applied Sciences, Intelligent Vehicles Lab (IVL), 80335 Munich, Germany
        {\tt\small intelligent-vehicles [at] hm.edu}}
}
\maketitle

\begin{abstract}
Vision-Language Models (VLMs) are increasingly used in autonomous-driving systems, yet their ability to recover dynamic physical state from visual input remains insufficiently characterized. We study velocity understanding as a controlled diagnostic across three tasks: surrounding-agent speed, current ego speed, and short-horizon future ego-speed proposal. On nuScenes, we evaluate open-weight general-purpose and Physical-AI VLMs, together with the driving-oriented Alpamayo-1.5 Vision-Language-Action model (VLA), using multiple input and output formulations. We combine verbal evaluation with temporal perturbations, counterfactual ego-speed hints and linear probes of hidden representations. The tasks exhibit distinct failure modes. Surrounding-agent speed is weakly encoded in an agent-specific form, whereas current ego speed is often internally accessible but poorly verbalized: continuous probes achieve 4.7--5.8~km/h Mean Absolute Error (MAE) compared with 10.2--16.8~km/h MAE for verbal outputs. Multiple frames provide inconsistent verbal gains to single frame inputs, and frame order is rarely exploited. Under non-optimized Quantized Low-Rank Adaptation (QLoRA), task-specific adaptation improves both task-relevant latent speed representations and verbal readout, but continuous surrounding-agent speed estimation remains weak, while most future-speed gains survive frame shuffling, indicating limited temporal grounding. Driving specialized Alpamayo-1.5 shows stronger latent representations for surrounding-agent and future ego speed, while current ego-speed decodability is comparable and substantial probe--verbal gaps remain. Thus, driving specialization can strengthen motion representations but does not guarantee stronger encoding across both scene and ego states or reliable readout. The results show that plausible planning outputs do not necessarily imply reliable recovery or temporal grounding of the underlying dynamic state.
\end{abstract}


\section{INTRODUCTION}
\label{sec:intro}

Vision-Language Models (VLMs) are increasingly used for perception, scene understanding, and behavior planning in autonomous-driving systems~\cite{zhou2024vision,drivegpt4,lingoqa,drivevlm}. A central open question is whether these models capture not only scene semantics, but also the dynamic state variables required for reliable prediction and action, including ego velocity and surrounding-agent motion.

Existing evaluations do not isolate this capability. General Video-VLM benchmarks report persistent weaknesses in motion and temporal understanding~\cite{mvbench,videomme,tempcompass,motionbench}, whereas driving benchmarks often combine perception with grounding, reasoning, and planning~\cite{fruhwirth2026stsbench,ishihara2026stride,zhou2025tumtraffic,lingoqa,drivelm,ding2024holistic}. Their aggregate scores therefore do not reveal whether an error arises from missing motion information, inadequate temporal integration, or failure to express an informative internal representation.

\begin{figure}[!t]
\centering
\includegraphics[width=\columnwidth]{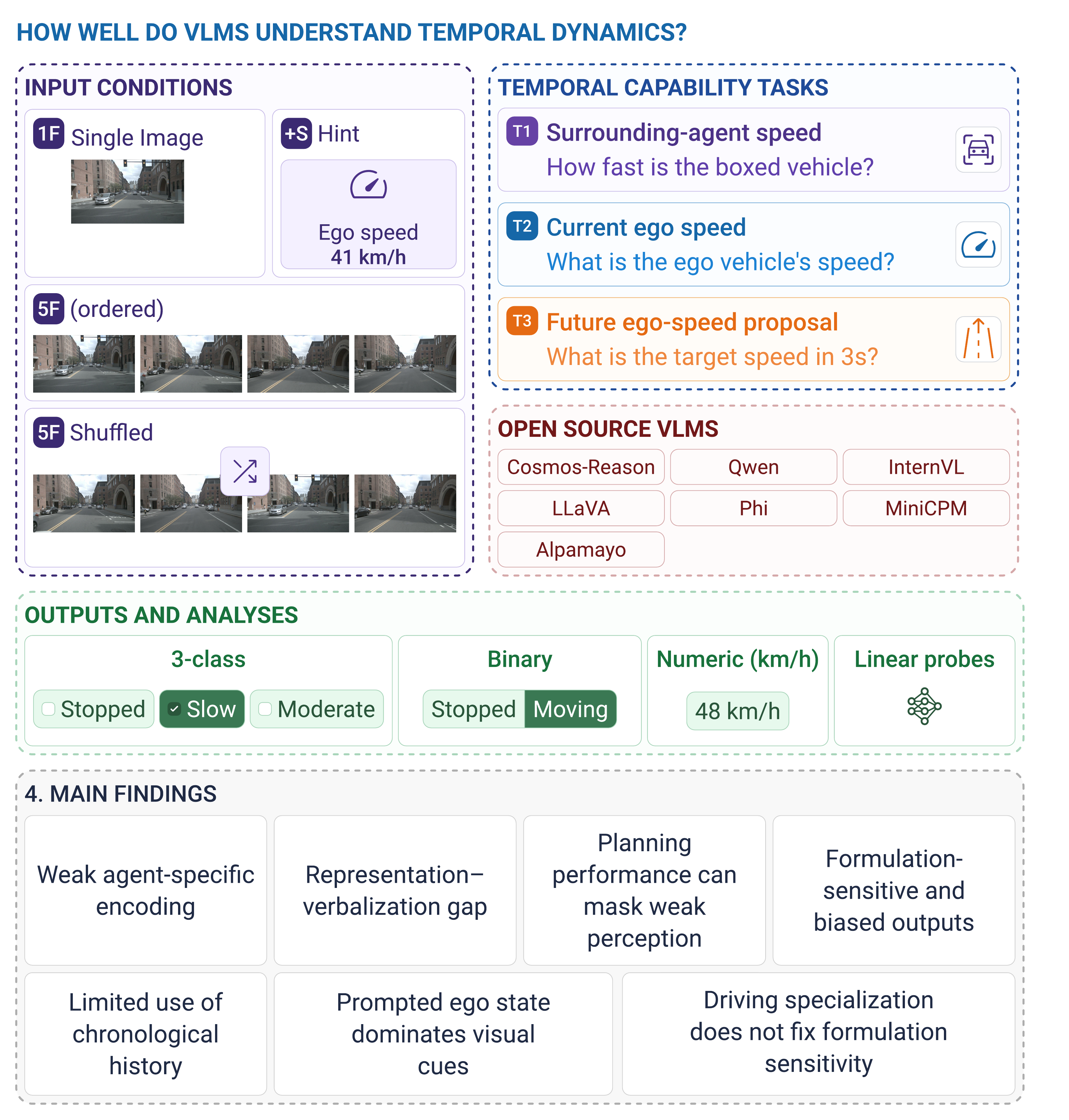}
\caption{Overview of the evaluation framework and main findings for temporal-dynamic understanding in driving VLMs. Across three velocity tasks and controlled visual and ego-state input conditions, we assess output behavior and hidden representations. The results reveal weak agent-specific encoding, representation--verbalization gaps, limited temporal-order use, strong reliance on supplied ego state, and only partial correction through task-specific adaptation.}
\label{fig:radar}
\end{figure}

We address this ambiguity by using velocity as a controlled diagnostic connecting dynamic perception and planning-related output. On nuScenes~\cite{nuscenes}, we evaluate three tasks: surrounding-agent speed (T1), current ego speed (T2), and short-horizon future ego speed (T3), under categorical and metric formulations. 
Rather than evaluating output accuracy alone, we ask whether the relevant state is encoded and reflected in the verbal output, whether models exploit temporal evidence and its chronological structure by comparing single-frame, ordered multi-frame, and temporally perturbed inputs, and how predictions respond when accurate or erroneous ego-state information is supplied.

This distinction is important for autonomous-driving research because a plausible future-speed proposal does not establish that the model has recovered the dynamic state on which that proposal should depend. An output may agree with a recorded trajectory even when current ego speed is poorly estimated from video or when chronological motion evidence contributes little to the prediction. Conversely, an incorrect output may conceal informative internal representations that are not faithfully exposed through the language interface.

Our main analysis evaluates open-weight general-purpose, video-capable, and Physical-AI VLMs, while a separate adaptation study includes Alpamayo-1.5 as a driving-oriented Vision-Language-Action model (VLA) reference~\cite{alpamayo15}. As summarized in Fig.~\ref{fig:radar}, our results reveal distinct failure stages. Surrounding-agent speed is only weakly encoded in an agent-specific form, whereas current ego speed is substantially more accessible in hidden representations than in verbal outputs. Correct ego-speed hints strongly improve both current- and future-speed reporting, demonstrating the utility of accurate current-state information in the prompt. However, counterfactual hints sharply degrade both tasks, indicating that the models do not robustly verify auxiliary prompted state against the video. Zero-shot predictions are largely insensitive to frame order and the addition of multi-frame history, but sensitive to output formulation and bias. Task-specific adaptation improves representations and reporting, but most of the future-speed gain remains under shuffled input, while only improving on the specific output formulation.

This paper makes three key contributions:
\begin{itemize}
    \item We introduce a controlled evaluation protocol for VLMs in driving scenarios, focusing on dynamic state recovery through three tasks: surrounding-agent speed estimation, current ego-speed detection, and 3-second-ahead ego-speed prediction. Our protocol tests models under single-frame, ordered/unordered sequences, and ego-state hint input conditions against chance, optimal-constant, and persistence baselines. Our results show that even the best models fail to outperform simple heuristics for future-speed prediction (6.5 km/h worse than persistence), and prompt-supplied ego state often overrides visual evidence.

    \item To dissect model failures, we combine linear probes, query-conditioned decoding, and controlled perturbations to separate encoding, readout and temporal grounding. We find that surrounding-agent speed is weakly encoded, whereas current ego-speed is more accessible internally than verbally, with regression probes achieving 4.7--5.8 km/h MAE versus 10.2--16.8 km/h for verbal outputs. Ordered history provides no consistent advantage over unordered frames, revealing gaps in temporal reasoning and output fidelity.

    \item Finally, we demonstrate that task-specific fine-tuning (e.g., Quantized Low Rank Adaptation (QLoRA)) can improve ego-speed readout dramatically (40.9 $\rightarrow$ 90.1\% macro-F1 for Qwen3-VL on ego-speed estimation), while temporal grounding for future speed proposal remains limited. Driving-specialized model Alpamayo-1.5 also exhibits a substantial probe--verbal mismatch, suggesting systemic limitations in dynamic state recovery. We summarize the main findings and corresponding mitigation directions in Table~\ref{tab:discussion_overview} and release our evaluation suite and probing toolkit to facilitate further research on our project page \url{https://iv.ee.hm.edu/projects/speed-in-the-blind-spot/}.

\end{itemize}

The remainder of the paper is organized as follows. Section~\ref{sec:related} discusses related work on driving VLMs, benchmarks, and interpretability, while Section~\ref{sec:setup} describes the tasks, data, prompting, models, and evaluation metrics. Section~\ref{sec:results} presents results on zero-shot dynamic-state estimation across tasks and input conditions, followed by representation probing and temporal-grounding analyses in Section~\ref{sec:representations}. Section~\ref{sec:lora} shows the analysis task-specific adaptation and residual failure modes. Finally, Section~\ref{sec:discussion} discusses the main findings and implications, and Section~\ref{sec:conclusion} concludes the paper.

\section{RELATED WORK}
\label{sec:related}

VLMs and VLAs have been integrated into autonomous-driving systems for tasks such as planning~\cite{drivelm,drivegpt4,dolphins}, scene understanding~\cite{lingoqa}, and behavior conditioning or action generation~\cite{drivevlm,talk2bev,alpamayo15}. Many of these tasks depend on an accurate representation of the current dynamic state, particularly ego and surrounding-agent velocity. Li et al.~\cite{li2024ego} show that explicit ego-state information can dominate open-loop driving performance, highlighting both its importance and the risk that models exploit it as a shortcut rather than infer it visually. At the same time, existing VLM driving tasks usually represent velocity coarsely or embed it within broader grounding, reasoning, and planning objectives~\cite{drivegpt4,drivevlm,dolphins,talk2bev}.

Video-capable VLMs incorporate temporal context through native video tokenization, temporal feature aggregation, or ordered multi-image input~\cite{llavavideo,qwen25vl,qwen3vl,internvl25,internvl2,minicpmv,phi35vision}. Across these architectures, benchmarks report persistent weaknesses in temporal ordering, motion direction, speed estimation, and future prediction, while also showing that apparent video understanding may rely on static cues, language priors, or task-format effects rather than ordered motion evidence~\cite{mvbench,tempcompass,motionbench,videomme,videohallucer,sicheng2025vitatecs,fang2024mmbench,onlytimecantell}.

Driving benchmarks further evaluate interaction understanding, motion status, prediction, and planning~\cite{lingoqa,drivelm,ding2024holistic,fruhwirth2026stsbench,ishihara2026stride,zhou2025tumtraffic}. VENUSS~\cite{brusnicki2026venuss} complements these benchmarks by systematically studying the sensitivity of sequential driving-scene understanding to visual input configuration. In contrast, we isolate metric, agent-specific velocity perception and examine how dynamic-state information is represented, temporally grounded, and exposed in model outputs.

Interpretability studies use linear probes to localize visual concepts in multimodal representations~\cite{probing-vlm} and extract interpretable control directions from motion-forecasting transformers~\cite{wordsinmotion}. Theodoridis et al.~\cite{probingvlm-driving} use counterfactual CARLA images and linear probes to study whether driving-relevant concepts are represented internally. Their analysis focuses on discrete spatial concepts from single images. In contrast, we study continuous, agent-specific velocity and its temporal grounding.  Layer-wise analyses in language models further show that task-relevant information may emerge in intermediate layers and become less accessible toward the output~\cite{tenney-bert,skean-layers}. Existing work has not established how continuous dynamic-state information is represented across the visual encoder and language backbone of driving VLMs, whether it is temporally grounded and faithfully reflected in model outputs, or how these properties change under task-specific adaptation.

\section{EXPERIMENTAL SETUP}
\label{sec:setup}

This section defines the three velocity tasks and describes the dataset, prompting conditions, evaluated models, and metrics used throughout the study. Together, these choices provide a controlled basis for comparing verbal performance, internal representations, and adaptation effects.

\subsection{TASKS}

We evaluate three complementary tasks: speed of a surrounding vehicle marked by a green bounding box (T1), current ego speed (T2), and proposed ego speed 3 seconds ahead (T3). T1 and T2 probe current-state perception but emphasize different evidence: T1 requires isolating the marked agent's motion relative to the ego vehicle and scene, whereas T2 relies primarily on ego-induced global scene motion. T3 requires integrating ego motion with scene context and, where applicable, surrounding-agent motion to predict future ego speed. T1 predicts the marked vehicle's speed class together with its vehicle type, which serves as an auxiliary consistency check for object identification. Together, the tasks span current-state perception and future-state proposal.

We evaluate two categorical granularities: a three-class formulation (\emph{stopped}, \emph{slow}, or \emph{moderate}) and a binary formulation (\emph{stopped} versus \emph{moving}). 
The threshold for \emph{stopped} is shared across both formulations and all tasks: \(v<1.8 km/h\); in the three-class setting, \emph{slow} corresponds to \(1.8\leq v\leq18 \mathrm{km/h}\) and \emph{moderate} to \(v>18 \mathrm{km/h}\). For the continuous formulation, models predict speed in km/h.

Following prior driving question-answering benchmarks~\cite{fruhwirth2026stsbench}, we formulate each task as closed-set question answering. We vary the visual context between a single frame (1F) and five frames (5F) sampled at 2Hz. The 1F condition serves as a non-temporal baseline, reflecting  what can be inferred from scene semantics, motion blur and priors, whereas 5F additionally provides explicit motion evidence. We independently vary whether the prompt includes a numerical ego-speed hint~(S). This yields the input conditions \textsc{1F}, \textsc{5F}, \textsc{1F+S}, and \textsc{5F+S}. As LLaVA-1.5 accepts only a single image, it is evaluated under \textsc{1F} and \textsc{1F+S} only.
To test whether models follow supplied state information or verify it against visual evidence, we additionally evaluate counterfactual hints in which the ego-speed value is replaced by a value from an incorrect speed class.

\begin{table}[!t]
\centering
\caption{%
Vision encoders and language-model backbones of the evaluated open source VLMs.
}
\label{tab:model_architectures}
\setlength{\tabcolsep}{0.1pt}
\renewcommand{\arraystretch}{1.00}
\footnotesize
\begin{tabular}{@{}lll@{}}
\toprule
\textbf{Model} & \textbf{Vision encoder} & \textbf{LLM backbone} \\
\midrule

Cosmos-Reason2-8B~\cite{cosmosreason2}
& SigLIP-2~\cite{tschannen2025siglip}
& Qwen3-8B \\

Cosmos-Reason1-7B~\cite{cosmosreason}
& Qwen2.5-VL ViT
& Qwen2.5-7B \\

Qwen2.5-VL-7B~\cite{qwen25vl}
& Qwen2.5-VL ViT
& Qwen2.5-7B \\

Qwen3-VL-8B~\cite{qwen3vl}
& SigLIP-2~\cite{tschannen2025siglip}
& Qwen3-8B \\

InternVL2-8B~\cite{internvl2}
& InternViT-300M-448px
& internlm2\_5-7b-chat \\

InternVL2.5-8B~\cite{internvl25}
& InternViT-300M-448px-V2.5
& internlm2\_5-7b-chat \\

LLaVA-Video-7B~\cite{llavavideo}
& SigLIP-SO400M
& Qwen2-7B \\

LLaVA-1.5-7B~\cite{llava15}
& CLIP ViT-L~\cite{radford2021learning}
& Vicuna-7B \\

Phi-3.5-Vision (4B)~\cite{phi35vision}
& CLIP ViT-L~\cite{radford2021learning}
& Phi-3.5 Mini \\

MiniCPM-V-2.6 (8B)~\cite{minicpmv}
& SigLIP-SO400M~\cite{zhai2023siglip}
& Qwen2-7B \\

\midrule
Alpamayo-1.5-10B~\cite{alpamayo15}
& SigLIP-2~\cite{tschannen2025siglip}
& Qwen3-8B \\

\bottomrule
\end{tabular}
\end{table}

\begin{table*}[!t]
\centering
\caption{Three-class macro-F1 (\%) for surrounding-agent speed (T1), current ego speed (T2), and 3s-ahead ego-speed proposal (T3) under single-frame (1F), five-frame (5F), and ego-speed-hint (+S) conditions. Visual-only performance remains strongly task- and model-dependent, with no consistent benefit from five-frame input. Ego-speed hints yield substantial gains for T2 and T3. Only MiniCPM-V$^\dagger$ emitted unparsable responses, for 3.9\% of samples on T1, 5F+S and $\leq$0.3\% for other conditions.}
\label{tab:main_results}
\setlength{\tabcolsep}{1.5pt}
\tiny
\resizebox{0.6\textwidth}{!}{%
\begin{tabular}{@{}l cccc cccc cccc@{}}
\toprule
& \multicolumn{4}{c}{T1 surrounding vehicle} & \multicolumn{4}{c}{T2 ego current} & \multicolumn{4}{c}{T3 ego future} \\
\cmidrule(lr){2-5}\cmidrule(lr){6-9}\cmidrule(lr){10-13}
Model & 1F & 5F & 1F+S & 5F+S & 1F & 5F & 1F+S & 5F+S & 1F & 5F & 1F+S & 5F+S\\
\midrule
Cosmos-R2 & 22.1 & 21.8 & 27.3 & 29.1 & 22.9 & 36.2 & 65.2 & 63.4 & 46.1 & 44.5 & \textbf{66.1} & 61.0 \\
Cosmos-R1 & \textbf{43.3} & \textbf{42.2} & \textbf{44.7} & \textbf{44.7} & 31.4 & 33.1 & 57.2 & 62.0 & 45.6 & \textbf{55.1} & 64.8 & 72.7 \\
Qwen2.5-VL & 33.5 & 39.1 & 41.6 & 43.2 & 22.7 & 27.8 & 50.2 & 64.8 & 35.3 & 35.3 & 52.9 & 54.3 \\
Qwen3-VL & 42.0 & 30.5 & 37.8 & 39.1 & \textbf{40.4} & 40.9 & \textbf{78.8} & \textbf{82.0} & \textbf{51.8} & 51.4 & 61.1 & 69.4 \\
LLaVA-Video & 21.6 & 14.2 & 18.4 & 20.3 & 23.0 & 25.3 & 37.2 & 47.4 & 14.6 & 14.1 & 15.2 & 14.2 \\
InternVL2.5 & 41.7 & 40.4 & 40.9 & 39.1 & 28.3 & 23.4 & 56.5 & 78.7 & 41.4 & 49.0 & 61.4 & \textbf{73.0} \\
InternVL2 & 38.6 & 39.4 & 37.4 & 39.5 & 23.2 & 13.8 & 32.5 & 26.7 & 43.2 & 38.2 & 54.0 & 68.9 \\
Phi-3.5-V & 31.5 & 26.2 & 38.7 & 35.9 & 15.3 & 16.2 & 52.0 & 56.1 & 31.3 & 19.5 & 42.0 & 30.1 \\
MiniCPM-V$^\dagger$ & 37.1 & 31.5 & 33.8 & 34.5 & 30.6 & \textbf{44.9} & 70.6 & 67.1 & 40.4 & 31.6 & 50.2 & 47.7 \\
LLaVA-1.5 & 30.4 & -- & 29.2 & -- & 23.9 & -- & 30.2 & -- & 25.5 & -- & 24.2 & -- \\

\bottomrule
\end{tabular}%
}
\end{table*}

\subsection{DATASET AND SPLITS}

We use the official nuScenes~\cite{nuscenes} training and validation splits, which remain disjoint throughout all experiments. Table~\ref{tab:data_splits} summarizes how the splits are used for benchmark evaluation, probing, and task-specific adaptation.

The validation split contains 6,019 keyframes from 150 scenes at 2Hz. Excluding the first four keyframes of each scene to provide five-frame history leaves 5,419 samples, which form the T2 evaluation pool and are also used for single-frame evaluation. T1 retains the 4,429 samples with a visible front-camera target vehicle, while T3 retains 4,519 samples with a valid 3s future trajectory. The corresponding \{stopped, slow, moderate\} class distributions are \{54.0, 14.0, 31.9\%\} for T1, \{16.6, 27.5, 56.0\%\} for T2, and \{16.3, 26.9, 56.8\%\} for T3.
All zero-shot, probing, and adaptation results are evaluated on these same validation pools. Training for probing and adaptation uses only scenes from the disjoint nuScenes training split. Linear probes use fixed 6,000-sample training subsets drawn across 547 scenes. The scene count therefore reflects broad coverage rather than dense sampling within individual scenes. Task-specific QLoRA adaptation instead uses the full eligible training split of 20,275 samples with at least four history frames, spanning 560 scenes. For each task, a weighted sampler draws samples inversely to the frequency of its target classes, balancing surrounding-agent, current-ego, or future-ego speed, respectively.

\begin{table}[!t]
\centering
\caption{%
Use of the nuScenes training and validation splits across the study. Training and validation scenes are disjoint. Sample counts are a subset of the original split reflecting valid temporal history, target visibility, and future trajectories, as well as balanced sampling for probing and adaptation.
}
\label{tab:data_splits}
\setlength{\tabcolsep}{2.5pt}
\renewcommand{\arraystretch}{1.00}
\footnotesize

\begin{tabular}{@{}llll@{}}
\toprule
\textbf{Purpose} &
\textbf{Split} &
\textbf{Scenes} &
\textbf{Samples} \\
\midrule

Benchmark evaluation
& Val.
& 150
& T1: 4,429; T2: 5,419; T3: 4,519 \\

Probe training
& Train
& 547
& 6,000 ($\approx$2,000/class;joint-stratified) \\

QLoRA adaptation
& Train
& 560
& 20,275 \\

\bottomrule
\end{tabular}
\end{table}

\subsection{PROMPTING CONDITIONS}
All experiments use a two-turn system--user prompting setup. The system message instructs the model to return valid JSON, while the user message contains the scene context, the task-specific question, the class definitions, and the required output schema. In the \textsc{+S} setting, the current ego speed in km/h is additionally provided in the user prompt.

For example, the T3 \textsc{5F+S} prompt is:
\begin{quote}
\footnotesize
You are observing 5 dashcam frames from the ego vehicle in chronological order
(oldest to most recent). The frames are captured 0.5 seconds apart, covering
approximately 2 seconds of driving history. The last image is the current moment.
Your current driving speed is \texttt{<X.X>} km/h.
Based on the traffic situation as of the last (current) frame and how the scene has evolved, what speed should the ego vehicle drive in the next few seconds? 
Speed classes for the next action: stop = come to a stop or remain stopped; slow = drive slowly, below city speed $\sim 2$--$18$ km/h); moderate = drive at city speed or above ($>18$ km/h). Respond with JSON only: {"speed\_class": "..."}.
\end{quote}

T1 instead requests the type and speed class of the green-boxed vehicle, while numerical variants return \texttt{speed\_kmh}. Outputs are parsed deterministically without an LLM judge. Alpamayo-1.5 receives equivalent A/B/C multiple-choice prompts, with each option mapped to one class, because it does not reliably follow the JSON schema. Full prompt templates are included in the released toolkit.

\begin{figure*}[!t]
\centering
\includegraphics[width=0.8\textwidth]{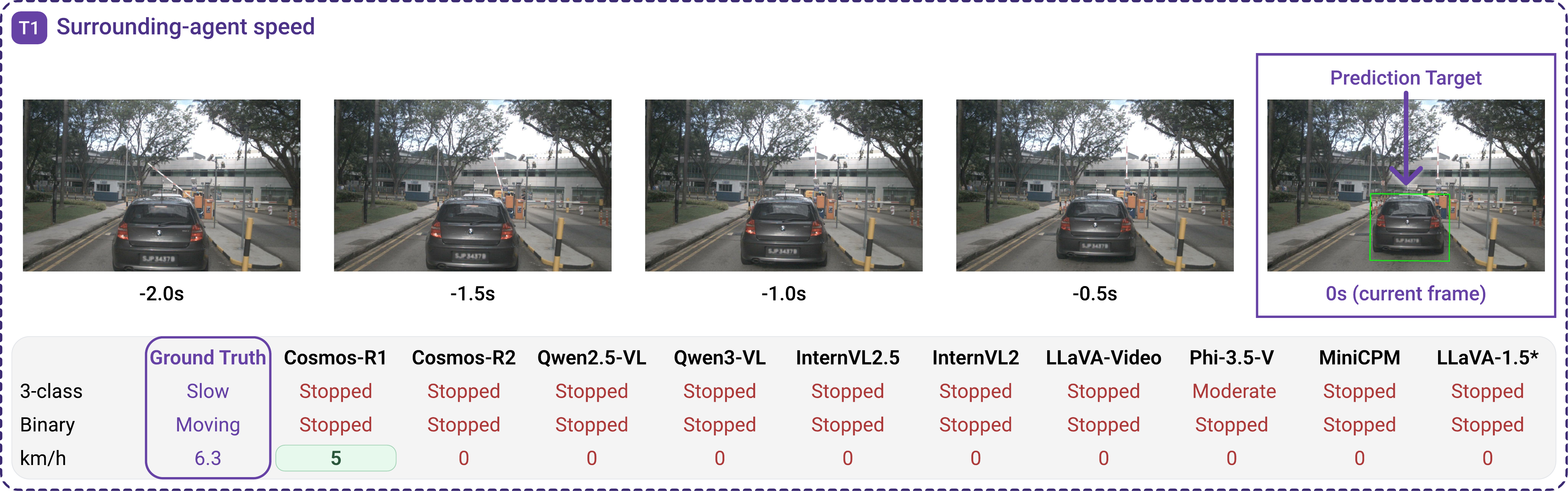}
\caption{Qualitative T1 failure case selected to illustrate errors in agent-speed estimation. The five input frames, ground-truth lead-vehicle speed, and corresponding model predictions are shown. Only Cosmos-R1 approximates the ground-truth speed in the regression output, with a 21\% relative error. $^\ast$LLaVA-1.5 receives only the current frame.}
\label{fig:qualitative-samples}
\end{figure*}

\subsection{MODELS}

Table~\ref{tab:model_architectures} summarizes the evaluated models and their visual and language backbones. The main benchmark includes only open-weight general-purpose and Physical-AI foundation models, thereby isolating dynamic-state capabilities acquired through broad multimodal pretraining and post-training before driving-specific adaptation. Open weights also enable hidden-state probing and controlled finetuning. Cosmos-Reason1/2 provide physically oriented variants of Qwen2.5-VL and Qwen3-VL, respectively.
For computationally intensive analyses beyond the full benchmark, we use a fixed six-model subset comprising Cosmos-Reason1/2, Qwen2.5-VL, Qwen3-VL, InternVL2.5, and LLaVA-Video. Alpamayo-1.5 is excluded from the main comparison because its prior adaptation to autonomous-driving data makes it not directly comparable to broadly trained foundation models. Instead, we treat it as a driving-adapted VLA reference, asking whether zero-shot and task-specific QLoRA show similar effects in a foundation VLM and a model already specialized for driving. This separates broadly acquired capabilities from those shaped by driving- and task-specific supervision.

Our implementation uses the native video interfaces of Cosmos-Reason1/2, Qwen2.5/3-VL, MiniCPM-V-2.6 and LLaVA-Video. InternVL2/2.5 and Phi-3.5-Vision receive chronologically ordered images, while LLaVA-1.5 is limited to a single frame. Alpamayo-1.5 likewise receives ordered multi-timestep images. Alpamayo-1.5 was trained on nuScenes-derived data, so its results may be affected by data contamination. For other models, nuScenes exposure is undocumented but cannot be ruled out because training data is only partly disclosed.
All models use greedy decoding (temperature 0; Alpamayo-1.5 near-greedy at 0.01), making evaluation deterministic to enable reproducibility. Zero-shot inference and feature extraction were performed on a single NVIDIA A40 48 GB or A100-SXM4 80 GB GPU, depending on model size.

\subsection{METRICS AND EVALUATION}\label{sec:metrics}
Categorical outputs are evaluated by macro-F1, the unweighted mean of the per-class F1 scores, where each F1 is the harmonic mean of precision and recall. As an input-independent reference, predictions are sampled from the empirical class distribution of each task; the expected macro-F1 is \(1/K\), i.e., \(33.3\%\) for three classes and \(50\%\) for binary tasks. Continuous predictions are evaluated by mean absolute error (MAE), the average absolute deviation from the ground-truth speed in km/h,  root mean squared error (RMSE) gives greater weight to large errors; and Spearman's \(\rho\), which measures rank-order agreement independently of absolute scale. 
Since continuous regression has no natural chance level, we report an optimal constant reference for MAE, obtained by predicting the median ground-truth speed of the corresponding evaluation pool for every sample. The resulting constant predictions are 0.7, 20.2, and 20.7 km/h for T1--T3, respectively. Because T1--T3 use different evaluation pools and target distributions, absolute metric values do not provide a direct measure of relative task difficulty. We therefore interpret three-class and binary macro-F1 relative to their respective stochastic references, and continuous MAE relative to the corresponding task-specific constant baseline.
Outputs are parsed deterministically using schema-field extraction and a fixed synonym map, while refusals and unparsable responses count as failures.

\section{ZERO-SHOT DYNAMIC-STATE RECOVERY}
\label{sec:results}

VLMs show persistent weaknesses in motion and temporal understanding. We explicitly study velocity as a controlled physical quantity to localize these failures.

\subsection{OVERALL SPEED ESTIMATION PERFORMANCE}
\label{sec:results_verbal}

Table~\ref{tab:main_results} reports three-class macro-F1 for surrounding-agent speed (T1), current ego speed (T2), and future ego-speed proposal (T3). Under visual-only input, the best scores are 43.3\% for T1 with single-frame Cosmos-Reason1, 44.9\% for T2 with five-frame MiniCPM-V, and 55.1\% for T3 with five-frame Cosmos-Reason1. The task-best T1, T2, and T3 results exceed the 33.3\% stochastic reference by 10.0, 11.6, and 21.8 pp, respectively. Because the task-specific evaluation pools differ, these scores should be interpreted within each task rather than as a comparison of task difficulty. No model consistently dominates across tasks or input conditions, and additional frames provide no systematic improvement over a single frame. A qualitative T1 failure case is shown in Fig.~\ref{fig:qualitative-samples}.

Because T1 requires identifying and tracking a marked surrounding vehicle, we test whether its weak performance due to target association rather than speed estimation. Marking the target in every frame changes mean macro-F1 by only $-0.6$ pp, with no gain above 1.6 pp, suggesting that loss of target association is not the primary bottleneck. Vehicle-type recognition reaches 74.2\% mean accuracy (40.8\% macro-F1) but remains strongly biased toward \emph{car}. Thus, while target-recognition errors may contribute, they do not appear sufficient to explain the weak T1 performance. Full results are reported in Appendix~\ref{app:t1-controls}, Tables~\ref{tab:t1-box-control} and~\ref{tab:t1-type-control}.

\subsection{RELIANCE ON SUPPLIED EGO STATE}
\label{sec:ego-speed-hint}

\begin{figure}[!t]
\centering
\includegraphics[width=\columnwidth]{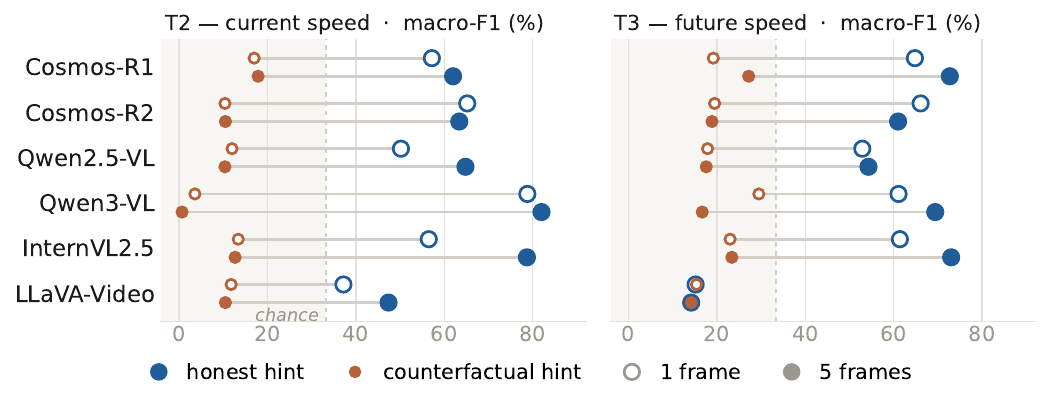}
\caption{Macro-F1 of correct and counterfactual ego-speed hints on T2 and T3. Counterfactual hints reduce performance below the stochastic reference, showing strong reliance on supplied ego state without reliable visual verification. Hollow and filled markers denote single-frame and five-frame input, respectively.}
\label{fig:false-hint}
\end{figure}

For T2, the ego-speed hint serves as a direct readout test. Because the supplied speed directly determines the target class, it separates failure to recover the current state visually from failure to use and report that state once it is given. Accordingly, the correct hint improves T2 for every model. For T3, current ego speed is auxiliary rather than the target, yet the hint also improves performance for all but LLaVA-1.5; Qwen3-VL reaches 82.0\% macro-F1 on T2 and InternVL2.5 73.0\% on T3. This shows that accurate current-state information is functionally useful for the planning-related future-speed output. For T1, effects are smaller and less consistent, with performance remaining at most 44.7\%. 
Additional analysis shows that, when ego speed is supplied, numerical predictions correlate substantially more strongly with ego speed than with the target vehicle's speed (Appendix~\ref{app:t1-ego-anchoring}, Table~\ref{tab:t1-ego-anchoring}), indicating anchoring on the supplied ego state rather than reliable integration with agent-specific motion cues.

We next test whether models verify supplied state against the visual input by replacing the correct hint randomly with a value from one of the two incorrect current-speed classes. Fig.~\ref{fig:false-hint} reports this intervention for the six-model analysis subset. Median macro-F1 falls below the 33.3\% stochastic reference for both T2 and T3, at 11\% and 19\%, respectively. Multi-frame input provides no benefit under counterfactual hints, except for Cosmos-R1 on T3, which remains below the reference.
Thus, VLM predictions are strongly influenced by supplied ego state and are not reliably corrected by conflicting visual evidence.

Together, the interventions show that this sensitivity to supplied ego state also extends to Physical-AI Foundation VLMs: accurate state improves current- and future-speed outputs, while erroneous state can strongly mislead them. This reliance suggests that planning-related T3 performance does not necessarily indicate reliable visual recovery or verification of the current dynamic state.

\begin{figure*}[!t]
\centering
\includegraphics[width=0.9\textwidth]{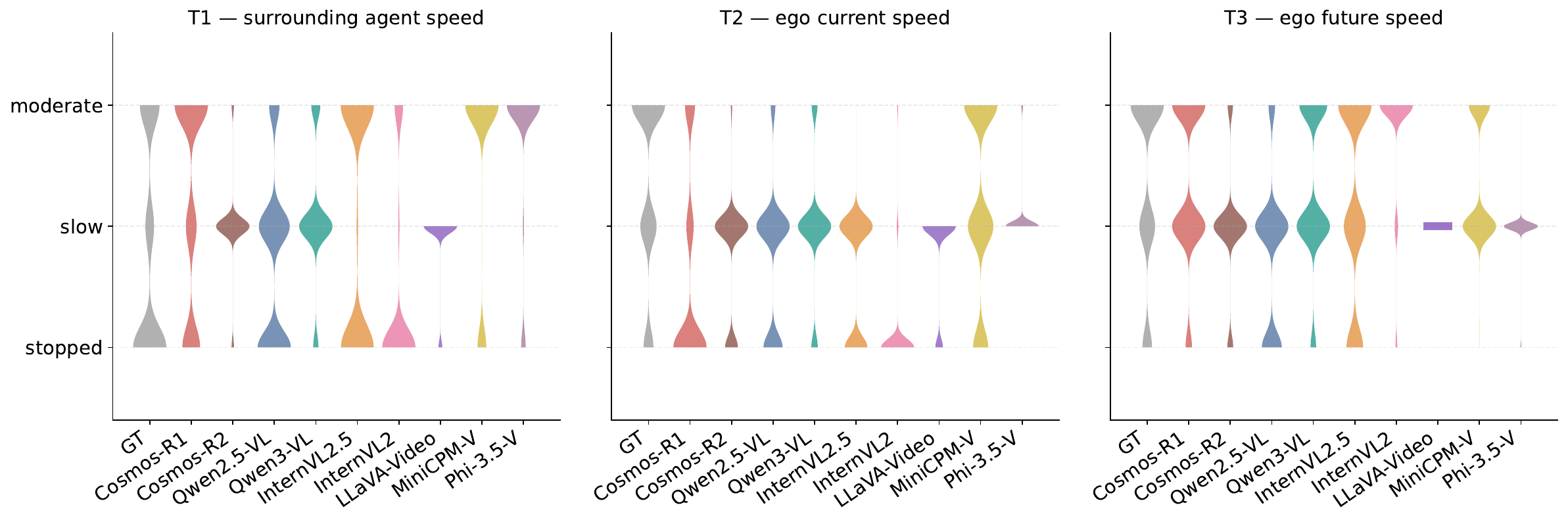}
\caption{Predicted class distributions for T1--T3 under five-frame visual-only input. GT denotes the empirical distribution and violin width prediction frequency. Biases vary across models and tasks.}
\label{fig:violin-class-dist}
\end{figure*}

\subsection{ROBUSTNESS TO TASK FORMULATION}
\label{sec:output_formulation}

\begin{table}[!t]
\centering
\caption{Binary stopped-versus-moving macro-F1 (\%) for the target surrounding vehicle (T1), the current ego state (T2), and the proposed ego state 3s ahead (T3). Results are reported for the visual-only single-frame (1F) and five-frame (5F) conditions. Reducing the task to stopped-versus-moving does not eliminate output collapse or the inconsistent benefit of additional frames.}
\label{tab:binary_f1}
\setlength{\tabcolsep}{4pt}
\footnotesize
\begin{tabular}{@{}l cc cc cc@{}}
\toprule
& \multicolumn{2}{c}{T1 surrounding}
& \multicolumn{2}{c}{T2 ego current}
& \multicolumn{2}{c}{T3 ego future} \\
\cmidrule(lr){2-3}\cmidrule(lr){4-5}\cmidrule(lr){6-7}
\textbf{Model} & 1F & 5F & 1F & 5F & 1F & 5F \\
\midrule
Cosmos-R2 &  60.1  &  \textbf{64.7}  &  25.9  &  \textbf{76.6}  &  46.5  &  \textbf{74.2}  \\
Cosmos-R1 &  35.7  &  35.2  &  16.3  &  15.8  &  32.8  &  22.6  \\
Qwen2.5-VL &  37.4  &  37.4  &  16.6  &  27.6  &  33.3  &  40.6  \\
Qwen3-VL &  \textbf{62.6}  &  46.2  &  \textbf{56.0}  &  67.9  &  59.3  &  67.8  \\
LLaVA-Video &  48.6  &  36.6  &  15.2  &  15.0  &  14.0  &  14.0  \\
InternVL2.5 &  62.2  &  61.7  &  37.1  &  40.0  &  \textbf{59.6}  &  69.6  \\
InternVL2 &  60.8  &  48.9  &  18.5  &  14.8  &  42.6  &  64.9  \\
Phi-3.5-V &  35.2  &  35.1  &  16.7  &  14.2  &  23.6  &  14.0  \\
MiniCPM-V &  35.1  &  35.5  &  14.2  &  14.2  &  14.2  &  14.0  \\
LLaVA-1.5 &  35.1  &  --  &  14.3  &  --  &  14.0  &  --  \\

\bottomrule
\end{tabular}
\end{table}

\begin{table}[!t]
\centering
\caption{Mean absolute error (MAE, km/h; lower is better) for numerical estimation of T1--T3 under 1F and 5F input. Continuous estimation remains challenging, with no consistent benefit from five-frame input. The optimal constant reference predicts the evaluation-set median for every sample. MAE is computed for parsable numerical outputs. Parsing rates are 99.8\% for Phi-3.5-V on T1--5F and 98.5\%, 77.5\%, and 82.6\% for MiniCPM-V$^\dagger$ on T2--1F, T2--5F, and T3--1F, respectively.}
\label{tab:kmh_mae}
\setlength{\tabcolsep}{4pt}
\footnotesize
\begin{tabular}{@{}l cc cc cc@{}}
\toprule
& \multicolumn{2}{c}{T1 surrounding}
& \multicolumn{2}{c}{T2 ego current}
& \multicolumn{2}{c}{T3 ego future} \\
\cmidrule(lr){2-3}\cmidrule(lr){4-5}\cmidrule(lr){6-7}
\textbf{Model} & 1F & 5F & 1F & 5F & 1F & 5F \\
\midrule
Cosmos-R2 &  \textbf{10.8}  &  \textbf{10.5}  &  19.0  &  14.2  &  15.2  &  12.0  \\
Cosmos-R1 &  17.6  &  15.4  &  \textbf{14.9}  &  \textbf{10.2}  &  13.4  &  14.3  \\
Qwen2.5-VL &  11.7  &  17.0  &  18.4  &  12.2  &  12.8  &  14.2  \\
Qwen3-VL &  15.6  &  15.5  &  19.0  &  10.4  &  15.5  &  \textbf{10.6}  \\
LLaVA-Video &  11.6  &  11.9  &  19.1  &  18.5  &  15.1  &  14.5  \\
InternVL2.5 &  11.5  &  11.6  &  15.6  &  16.8  &  \textbf{12.2}  &  13.1  \\
InternVL2 &  13.6  &  15.4  &  15.9  &  14.1  &  17.9  &  13.1  \\
Phi-3.5-V &  11.8  &  12.0  &  19.0  &  18.9  &  19.3  &  21.1  \\
MiniCPM-V$^\dagger$  &  11.8  &  12.0  &  18.6  &  16.8  &  13.8  &  11.7  \\
LLaVA-1.5 &  10.9  &  --   &  17.8  &  --  &  \textbf{12.2}  & --  \\
\midrule
Optimal constant ref. & \multicolumn{2}{c}{ 11.7 } & \multicolumn{2}{c}{ 10.5 } & \multicolumn{2}{c}{ 10.7 } \\
\bottomrule
\end{tabular}
\end{table}

To test whether the main findings are specific to the three-class formulation, we repeat the evaluation with binary stopped-versus-moving classification and direct km/h prediction. In the binary task, Cosmos-Reason2 achieves the best five-frame score on all three tasks: 64.7\% for T1, 76.6\% for T2, and 74.2\% for T3, compared with the 50\% stochastic reference. Across the compared models, the coarser formulation does not eliminate output collapse, and additional frames again provide no consistent benefit.
The lowest MAEs are 10.5, 10.2, and 10.6 km/h for T1--T3 under multi-frame inputs. The evaluation-set median, which minimizes MAE among all constant predictions, already yields 11.7, 10.5, and 10.7 km/h for T1--T3. Thus, the best VLM results improve on their task-specific constant references by only 1.2, 0.3, and 0.1 km/h, respectively, indicating that the visual input provides only a small aggregate MAE advantage over predicting a single dataset-level speed.
For T3, a current-state persistence reference, $\hat{v}_{t+3\mathrm{s}}=v_t$, achieves an MAE of 4.1 km/h, substantially below the best VLM result of 10.6 km/h. Thus, much of the future-speed target is predictable from the current state alone. The remaining VLM error may reflect weak recovery of current ego speed, limited use of this persistence, or both.
Complementary RMSE results (Appendix~\ref{app:metric-results}) preserve the overall pattern while revealing differences in the tails of the error distributions. For T2 with 5F, Qwen3-VL has a slightly higher MAE than Cosmos-R1 (10.4 vs. 10.2 km/h) but a lower RMSE (13.7 vs. 15.5 km/h), indicating fewer large prediction errors. Individual samples also show cross-format inconsistencies, like classifying a vehicle as \emph{stopped} while predicting a clearly nonzero speed in the km/h formulation (Fig.~\ref{fig:qualitative-samples}).
To quantify these consistencies, we directly compare paired three-class and continuous predictions for identical zero-shot, no-hint inputs. Defining hard contradictions as non-adjacent class conflicts (\emph{stopped} with $\geq18$ km/h or \emph{moderate} with $<1.8$ km/h), we find them in 18.4\% of single-frame and 15.6\% of five-frame predictions. In addition, \emph{stopped} labels coincide with nonzero numerical speeds ($\geq1.8$ km/h) in 4.6\% and 8.6\% of cases, respectively. Thus, categorical and metric outputs do not consistently reflect the same speed estimate.

Overall, changing the output formulation affects absolute performance but does not remove the main failure modes: additional frames remain inconsistently useful, output collapse persists, and categorical and continuous predictions can disagree.

\subsection{PREDICTION BIASES AND OUTPUT COLLAPSE}
\label{sec:prediction_biases}

Fig.~\ref{fig:violin-class-dist} reveals strong but inconsistent prediction biases across both models and tasks. While several model--task combinations concentrate on a single class or approach output collapse, the preferred class varies substantially rather than reflecting one shared response bias. Overall, the errors reflect heterogeneous model- and task-specific biases rather than class imbalance alone.

\subsection{LIMITED SENSITIVITY TO FRAME ORDER}
\label{sec:frame_order}

\begin{figure}[!t]
\centering
\includegraphics[width=\columnwidth]{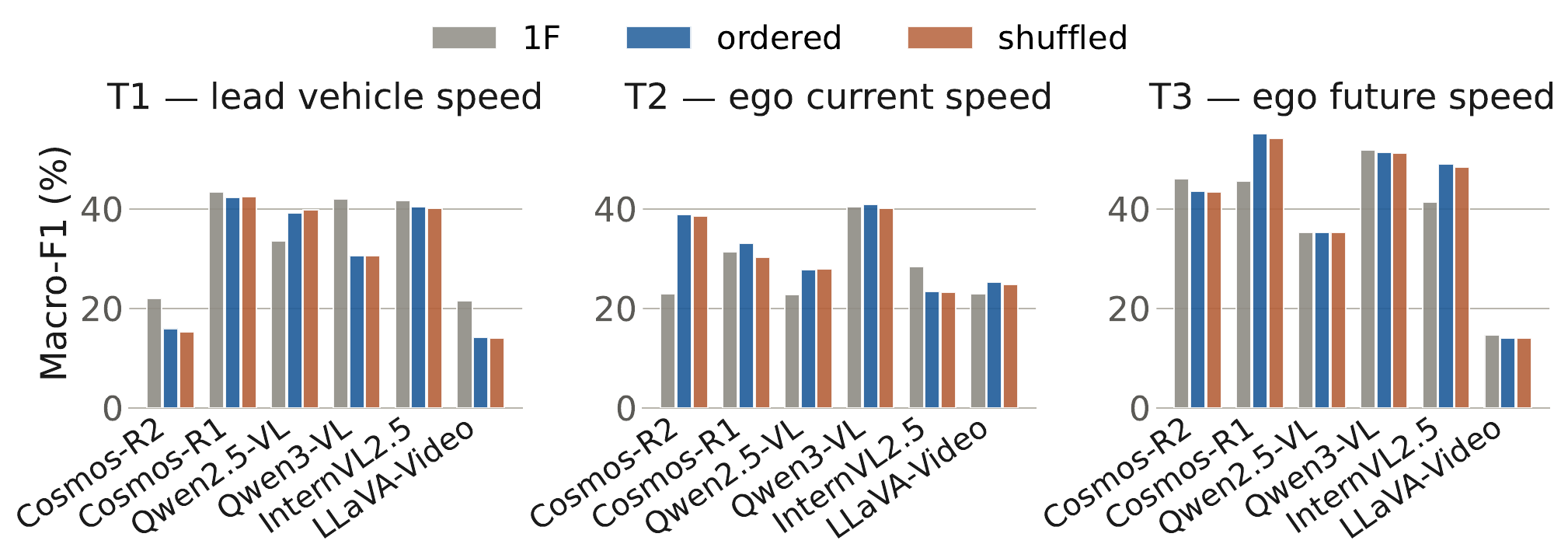}
\caption{Macro-F1 for single-frame, ordered five-frame, and shuffled history input with fixed current frame on the main evaluation pools. Additional frames are neutral or detrimental in most settings; where gains occur, they are often preserved under shuffling, consistent with a \emph{bag-of-frames} use of temporal context rather than strong reliance on chronological order.}
\label{fig:shuffle}
\end{figure}

To separate the contribution of additional visual observations from the use of their chronological order, we compare single-frame input with ordered and history-shuffled five-frame sequences for the six-model analysis subset. The current frame is kept fixed, while only the preceding history is shuffled.  Results are shown in Fig.~\ref{fig:shuffle}. Additional frames help in some model--task combinations, but gains are inconsistent.
Ordered and shuffled five-frame inputs usually perform similarly, suggesting a \emph{bag of frames} use of temporal context rather than reliance on chronological structure.

\begin{table*}[!t]
\centering
\caption{Last-layer probe macro-F1 (\%) for T1--T3. Parentheses show probe minus verbal performance in Table~\ref{tab:main_results} (pp). T2 and T3 are more strongly decodable from hidden representations than reflected in verbal outputs, whereas T1 shows no consistent probe advantage.}
\label{tab:probe_main}
\setlength{\tabcolsep}{5pt}
\footnotesize
\begin{tabular}{@{}l ccc | ccc | ccc@{}}
\toprule
 & \multicolumn{3}{c|}{\textbf{T1} -- lead speed}
 & \multicolumn{3}{c|}{\textbf{T2} -- ego speed}
 & \multicolumn{3}{c}{\textbf{T3} -- future speed} \\
\cmidrule(lr){2-4}\cmidrule(lr){5-7}\cmidrule(lr){8-10}
\textbf{Model} & 1F & 5F & 1F+S & 1F & 5F & 1F+S & 1F & 5F & 1F+S \\
\midrule
Cosmos-R1   & 35.6 ({\small-8})  & 49.7 ({\small+8})  & 43.4 ({\small-1})  & 59.6 ({\small+28}) & 82.3 ({\small+49}) & 96.6 ({\small+39}) & 57.4 ({\small+12}) & 66.2 ({\small+11}) & 69.2 ({\small+4}) \\
Cosmos-R2   & 36.7 ({\small+15}) & 44.4 ({\small+23}) & 38.8 ({\small+11}) & 62.3 ({\small+39}) & 75.3 ({\small+39}) & 95.4 ({\small+30}) & 59.5 ({\small+13}) & 64.6 ({\small+20}) & 68.6 ({\small+3}) \\
Qwen2.5-VL  & 38.9 ({\small+5})  & 29.1 ({\small-10}) & 41.9 ({\small+0})  & 60.5 ({\small+38}) & 77.4 ({\small+50}) & 96.9 ({\small+47}) & 56.5 ({\small+21}) & 67.5 ({\small+32}) & 73.3 ({\small+20}) \\
Qwen3-VL    & 25.6 ({\small-16}) & 41.1 ({\small+11}) & 21.8 ({\small-16}) & 61.3 ({\small+21}) & 77.3 ({\small+36}) & 96.9 ({\small+18}) & 59.3 ({\small+7}) & 61.8 ({\small+10}) & 68.5 ({\small+7}) \\
LLaVA-Video & 38.1 ({\small+17}) & 42.5 ({\small+28}) & 41.9 ({\small+24}) & 55.8 ({\small+33}) & 74.1 ({\small+49}) & 96.9 ({\small+60}) & 54.1 ({\small+40}) & 64.8 ({\small+51}) & 73.1 ({\small+58}) \\
InternVL2.5 & 27.9 ({\small-14}) & 33.4 ({\small-7})  & 42.3 ({\small+1})  & 57.7 ({\small+29}) & 75.9 ({\small+52}) & 97.1 ({\small+41}) & 56.4 ({\small+15}) & 63.6 ({\small+15}) & 72.2 ({\small+11}) \\
\bottomrule
\end{tabular}
\end{table*}

\section{DYNAMIC-STATE REPRESENTATIONS AND TEMPORAL GROUNDING}
\label{sec:representations}

Having characterized the models' verbal behavior, we next examine whether task-relevant speed information is accessible in their hidden representations. Due to the computational costs, we restrict the analysis to our six-model subset, unless stated otherwise.

The VLM parameters remain frozen throughout probing. For each task, input condition, and analyzed layer, we fit an independent linear probe on the extracted hidden representation. No probe gradients are propagated into the VLM. Vision-block outputs are mean-pooled over patch tokens, while LLM layers use the hidden state of the final prompt token.
Categorical probes are trained on the task-specific, joint-stratified nuScenes training subsets summarized in Table~\ref{tab:data_splits}, with balanced class weighting to account for residual class imbalance, and evaluated on the disjoint main evaluation pools. Features are standardized and reduced to 128 dimensions by PCA fitted on the training data, followed by multinomial logistic regression ($\ell_2$, $C{=}1$, L-BFGS, tolerance $10^{-4}$, maximum 1,000 iterations, seed 42). Probe metrics follow \subsecref{sec:setup}{sec:metrics}.

Table~\ref{tab:probe_main} compares probe and verbal performance under 1F, 5F, and 1F+S input on three-class labels. The subsequent layerwise, continuous, and cross-query analyses focus on five-frame input unless stated otherwise. Continuous probes use the same extracted representations and preprocessing but replace the classifier with ridge regression ($\alpha{=}10$). Both probe types are convex models optimized to convergence without an epoch schedule.

\subsection{REPRESENTATION-VERBALIZATION GAP}\label{sec:probe_main}

Table~\ref{tab:probe_main} shows a clear representation--verbalization gap for T2. Under visual-only input, probe macro-F1 rises from 55.8--62.3\% with 1F to 74.1--82.3\% with 5F, while verbal performance remains substantially lower, yielding gaps of up to 52 pp. Supplying the correct ego speed further raises 1F probe performance to 95.4--97.1\%, showing that the provided state is readily incorporated into the representation despite incomplete verbal readout.

T3 shows a smaller but consistent probe advantage. Visual-only probe performance rises from 54.1--59.5\% with 1F to 61.8--67.5\% with 5F, while a correct ego-speed hint raises 1F performance further to 68.5--73.3\%. Given that current ego speed alone is strongly predictive of the 3s-ahead target, however, T3 decodability may partly reflect current-state or scene information rather than a distinct representation of future ego speed. We therefore interpret T3 probes as evidence of target-predictive information, not by themselves as evidence of future-state reasoning or temporal grounding.

T1 shows no consistent probe advantage overall, only Cosmos-R2 and LLaVA-Video outperform their verbal outputs across all conditions. Five-frame input improves probe macro-F1 for five of six models (except Qwen2.5-VL), but scores remain close to or moderately above the 33.3\% chance reference; ego-speed hints provide no consistent benefit. This points to an encoding or agent-specific representation limitation rather than a pure verbalization failure.

\subsection{LAYERWISE LOCALIZATION OF SPEED INFORMATION}
\label{sec:layerwise}

\begin{figure}[!t]
\centering
\includegraphics[width=\columnwidth]{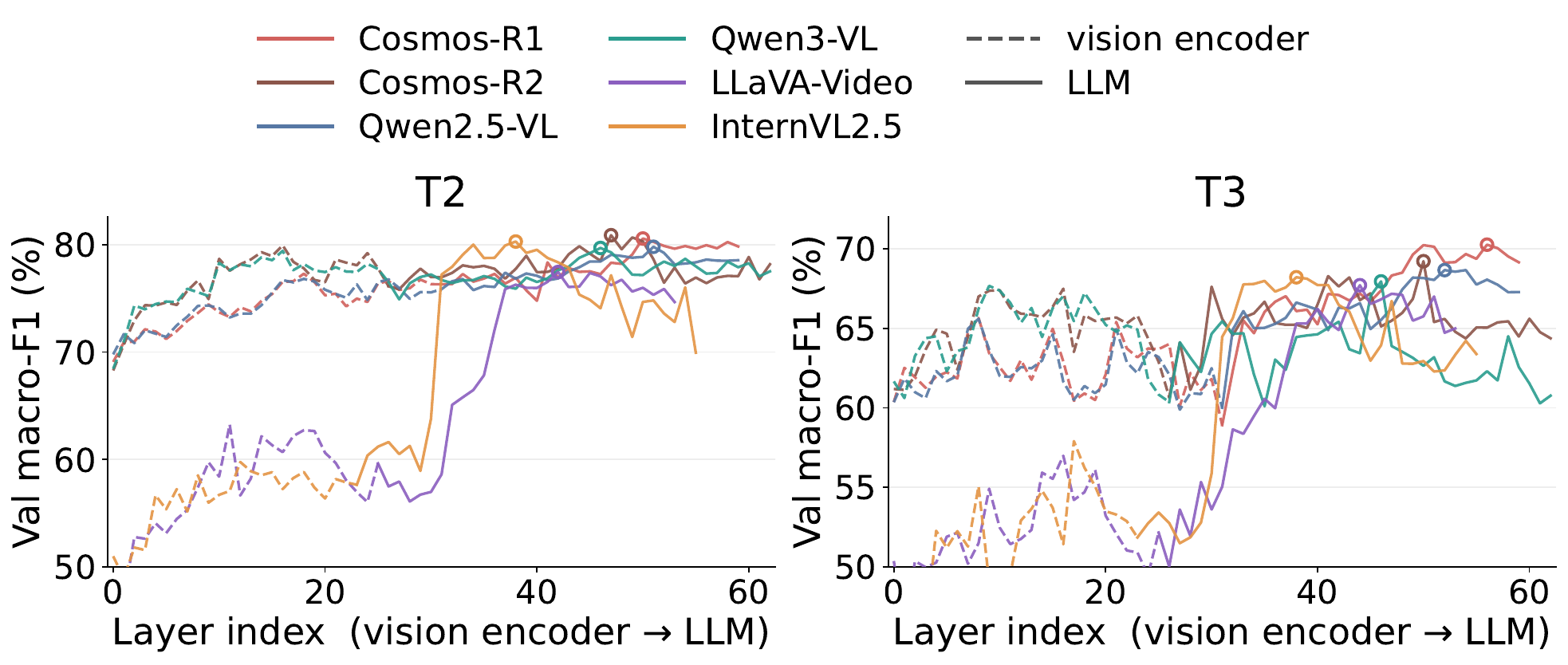}
\caption{Layerwise probe macro-F1 for T2 and T3 under five-frame input. Dashed and solid lines denote vision-encoder and LLM representations, respectively; circles mark peak layers. Current ego speed can already be strongly represented in visual features for some models, while future-speed-related information tends to emerge later in the LLM; in both cases, peak decodability often precedes the final layer.}
\label{fig:layerwise-probes}
\end{figure}

Fig.~\ref{fig:layerwise-probes} shows layerwise probes for the six models on T2 and T3. For the Cosmos and Qwen models, both tasks are already substantially decodable at the vision-encoder exit, whereas InternVL2.5 and LLaVA-Video gain more within the LLM. Peak decodability often occurs before the final layer, showing that task-relevant information is not preserved monotonically toward the output. These differences are not explained by encoder family alone: despite related SigLIP-family encoders, T2 becomes strongly decodable within the vision tower for Qwen3-VL but mainly within the LLM for LLaVA-Video, suggesting that projectors, language backbones, preprocessing, or training also shape its localization.

For T3, increasing decodability within the LLM may indicate additional processing beyond visual-state extraction. However, linear probes cannot distinguish future-motion reasoning from correlated cues such as current-speed persistence or scene priors.

\subsection{CONTINUOUS SPEED DECODING}\label{sec:kmh-scatter}

\begin{figure}[!t]
\centering
\includegraphics[width=\columnwidth]{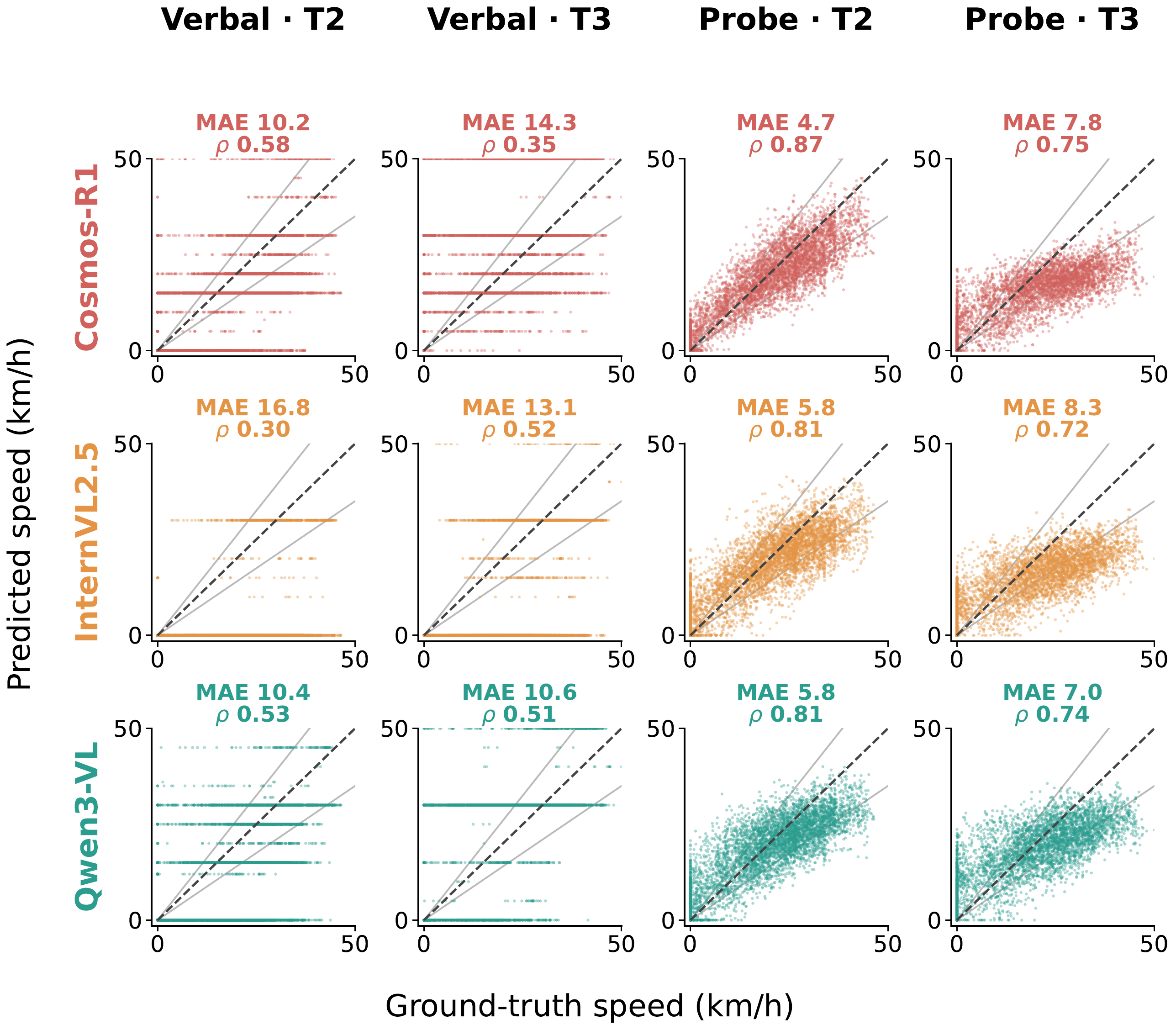}
\caption{Ground-truth versus predicted speed for numerical verbal outputs and last-layer regression probes under five-frame input. Panels report MAE and Spearman's $\rho$; the dashed line denotes perfect agreement and gray lines the $\pm30\%$ range. Probes recover substantially more precise speed information than verbal outputs, particularly for T2.}
\label{fig:kmh-scatter}
\end{figure}

Fig.~\ref{fig:kmh-scatter} shows that the representation--verbalization gap extends to continuous speed estimation for the representational models Cosmos-R1, InternVL2.5 and Qwen3-VL. For T2, last-layer regression probes achieve 4.7--5.8~km/h MAE with $\rho=0.81$--$0.87$, compared with 10.2--16.8~km/h and $\rho=0.30$--$0.58$ for verbal estimates. Thus, substantially more accurate current-speed information is linearly accessible in the hidden states than is expressed numerically.
For T3, the gap is smaller but remains consistent: probes achieve 7.0--8.3~km/h MAE with $\rho=0.72$--$0.75$, compared with 10.6--14.3~km/h and $\rho=0.35$--$0.52$ for verbal outputs. Probe predictions nevertheless exhibit a compressed range and predict lower speeds at higher target speeds. Thus, more accurate target-predictive speed information is linearly accessible in the hidden states than is expressed verbally, although T3 decodability alone does not establish reliable future-motion reasoning or temporal grounding. 

\subsection{EGO-AGENT REPRESENTATIONAL ASYMMETRY}\label{sec:probe_crosstask}

\begin{figure}[!t]
  \centering
  \includegraphics[width=0.8\linewidth]{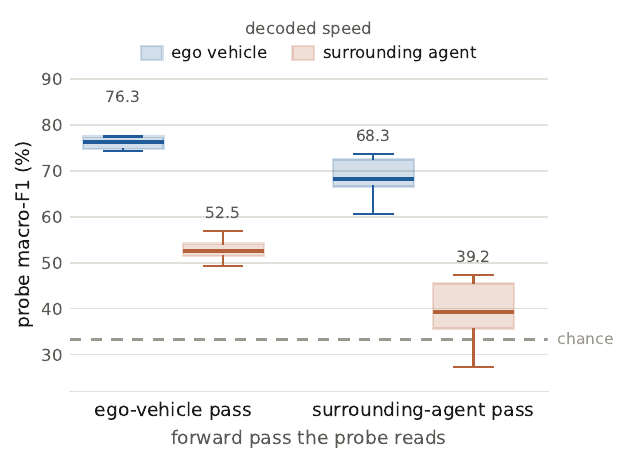}
\caption{Query-conditioned decoding of ego and surrounding-agent speed under five-frame input across six models. Probes decode both quantities from representations produced by either an ego-speed or agent-speed query. The ego-query representation is consistently more informative for both targets, while explicitly querying the marked vehicle does not improve agent-speed decodability.}
  \label{fig:probe_crosstask}
\end{figure}

To test whether asking about a specific motion state makes that state more accessible internally, we obtain two representations for each sample: one after querying the ego speed and one after querying the speed of the marked surrounding vehicle. From each representation, we then decode both quantities---ego speed and agent speed---using the same train-to-validation probing protocol as above. This yields a \(2\times2\) comparison between the queried quantity and the quantity decoded from the resulting representation shown in Fig.~\ref{fig:probe_crosstask}.
Taking the medians across the six models, the ego-query representation yields 76.3\% macro-F1 for ego speed and 52.5\% for agent speed, while the agent-query representation yields 68.3\% and 39.2\%, respectively. 
Thus, the ego-query representation is more informative for both quantities, and explicitly querying the surrounding vehicle does not improve decoding of its speed. This suggests that motion information is organized predominantly around the ego state rather than being restructured toward the queried agent. One possible explanation is that scene-wide ego-motion cues provide a useful reference for both ego and surrounding-agent motion, whereas the agent-specific query does not induce an equally informative agent-centric representation.

\section{TASK-SPECIFIC ADAPTATION AND RESIDUAL FAILURE MODES}
\label{sec:lora}

We next test whether the identified limitations can be reduced through task-specific adaptation. Qwen3-VL and the already driving-specialized Alpamayo-1.5 are independently fine-tuned for T1--T3 on the nuScenes training split (see Table~\ref{tab:data_splits}), with a class-balanced sampler. For Alpamayo-1.5, only the vision--language output pathway is used and further adapted, the action expert is not involved. The adapted models are evaluated on the same task-specific validation pools used throughout the preceding analyses.  Qwen3-VL retains the JSON-based prompts used in the main benchmark, whereas Alpamayo-1.5 uses equivalent A/B/C prompts. Note that \emph{zero-shot} denotes evaluation before any task-specific adaptation in this study. It does not imply absence of prior exposure to nuScenes or related driving data.

All runs use 4-bit NF4 QLoRA with bf16 adapters ($r{=}32$, $\alpha{=}32$, dropout $0.05$). Both models use AdamW with learning rate $2{\times}10^{-4}$, cosine decay, and 5\% warmup. Qwen3-VL is trained on four NVIDIA A100-SXM4 80 GB GPUs for 1,500 optimizer steps with effective batch size 32 and gradient clipping. Alpamayo-1.5 is trained on a single NVIDIA A40 48 GB GPU with effective batch size 8 using its model-specific, single-GPU training harness. Because the adaptation settings were not optimized for each model, we interpret these experiments diagnostically as demonstrating attainable changes rather than model-specific optima or upper bounds. We evaluate full vision--language adaptation for all tasks and restrict LLM-only and vision-encoder-only variants to T2. 

\begin{table}[!t]
\centering
\caption{Probe and verbal three-class macro-F1 (\%) before and after task-specific QLoRA, with adapter-placement ablations on T2. Adaptation improves both representation and readout in several settings, but not necessarily to the same extent. For T2, LLM-only adaptation is competitive with full adaptation. Training uses target-balanced weighted sampling, and probes use last-layer five-frame representations.}
\label{tab:ft-results}
\setlength{\tabcolsep}{5pt}
\footnotesize
\begin{tabular}{@{}ll l cc@{}}
\toprule
\textbf{Model} & \textbf{Task} & \textbf{Setting}
& \textbf{Probe} & \textbf{Verbal} \\
\midrule
\multirow{8}{*}{Qwen3-VL}
& T1 & Zero-shot     & 41.1 & 30.5 \\
&    & LoRA-full     & 69.9 & 71.0 \\
\cmidrule(l){2-5}
& T2 & Zero-shot     & 77.3 & 40.9 \\
&    & LoRA-full     & 89.0 & 90.1 \\
&    & LoRA LLM-only & 87.5 & 89.0 \\
&    & LoRA VE-only  & 85.9 & 87.6 \\
\cmidrule(l){2-5}
& T3 & Zero-shot     & 61.8 & 51.4 \\
&    & LoRA-full     & 71.0 & 72.4 \\
\midrule
\multirow{8}{*}{Alpamayo-1.5}
& T1 & Zero-shot     & 60.7 & 29.7 \\
&    & LoRA-full     & 63.6 & 52.9 \\
\cmidrule(l){2-5}
& T2 & Zero-shot     & 74.4 & 44.3 \\
&    & LoRA-full     & 83.0 & 79.4 \\
&    & LoRA LLM-only & 82.5 & 82.4 \\
&    & LoRA VE-only  & 67.4 & 47.5 \\
\cmidrule(l){2-5}
& T3 & Zero-shot     & 69.0 & 32.9 \\
&    & LoRA-full     & 77.0 & 50.6 \\
\bottomrule
\end{tabular}
\end{table}

\begin{figure*}[!t]
\centering
\includegraphics[width=0.9\textwidth]{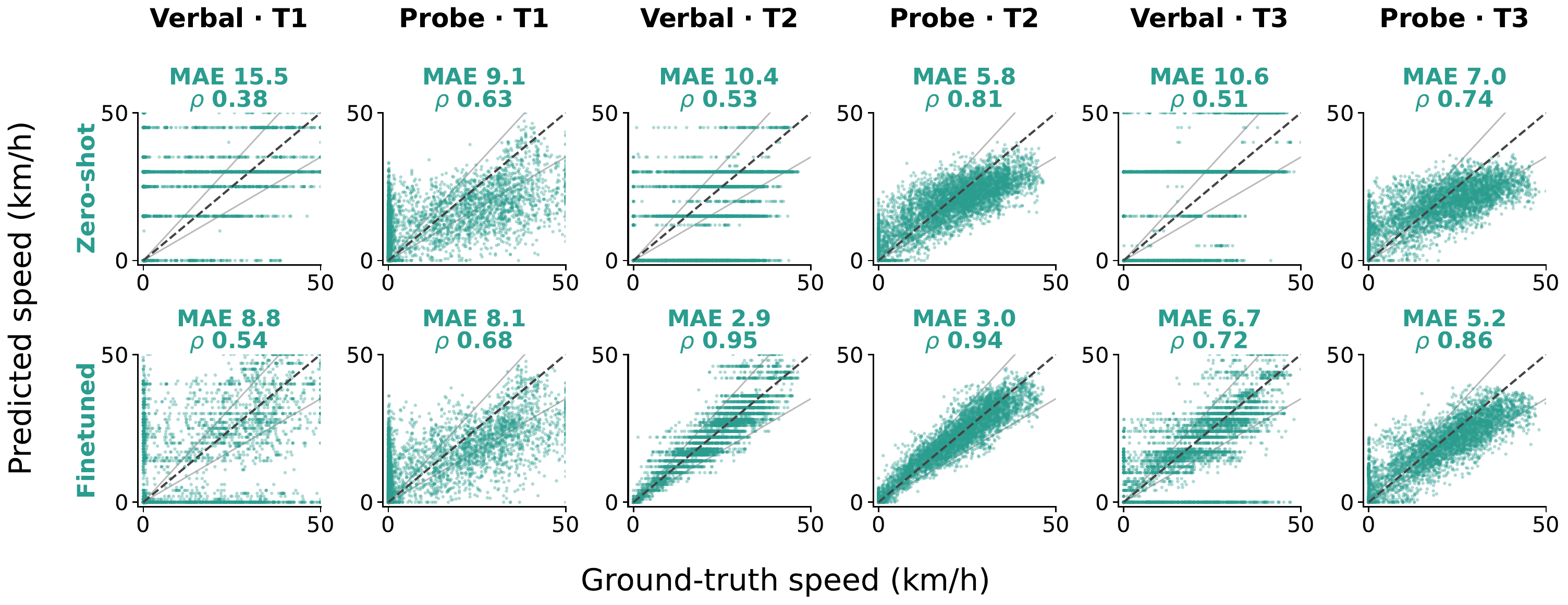}
\caption{Continuous speed estimation before and after task-specific full-LoRA adaptation of Qwen3-VL under five-frame input. Columns show verbal outputs and last-layer regression probes for T1--T3; rows show zero-shot and adapted models. The dashed line denotes perfect agreement and gray lines the $\pm30\%$ relative-error range. Adaptation nearly closes the continuous representation--verbalization gap for T2 and strongly improves T3, while T1 remains substantially harder even at the representation level.}
\label{fig:kmh_scatter_ft}
\end{figure*}

\begin{table}[!t]
\centering
\caption{%
Verbal and probe three-class macro-F1 (\%) for Qwen3-VL with ordered and randomly shuffled history inputs and fixed current frame before and after task-specific full-LoRA adaptation. T2 shows a larger ordered--shuffled gap after adaptation. For T3, adaptation improves overall performance while the ordered--shuffled gap remains small or decreases, indicating that most of the future-speed gain does not depend on chronological frame order.
}
\label{tab:shuffle_ft}
\setlength{\tabcolsep}{3.5pt}
\renewcommand{\arraystretch}{1.05}
\footnotesize

\begin{tabular}{@{}llcccc@{}}
\toprule
& &
\multicolumn{2}{c}{\textbf{Zero-shot}} &
\multicolumn{2}{c}{\textbf{LoRA-full}} \\
\cmidrule(lr){3-4}
\cmidrule(l){5-6}

\textbf{Task} &
\textbf{Readout} &
\textbf{Ordered} &
\textbf{Shuffled} &
\textbf{Ordered} &
\textbf{Shuffled} \\
\midrule

T2 & Verbal & 40.9 & 40.1 & 89.5 & 86.4 \\
   & Probe  & 77.3 & 76.5 & 89.0 & 83.6 \\
\midrule
T3 & Verbal & 51.4 & 51.3 & 72.4 & 72.2 \\
   & Probe  & 66.5 & 61.5 & 71.0 & 69.2 \\
   
\bottomrule
\end{tabular}
\end{table}

Table~\ref{tab:ft-results} shows that simple LoRA adaptation improves performance across all three tasks. Before task-specific adaptation, Alpamayo-1.5 exhibits substantially stronger latent T1 information than Qwen3-VL (60.7\% versus 41.1\% probe macro-F1), despite similarly weak verbal performance (29.7\% versus 30.5\%). This illustrates that similar output performance can mask substantial differences in representational strength. Because the models differ in architecture, output interface, and prior training exposure, this comparison is descriptive rather than a controlled estimate of the effect of driving specialization. On T2, Qwen3-VL and Alpamayo-1.5 show broadly similar probe decodability (77.3\% and 74.7\%), while verbal performance remains much lower for both. Driving-oriented training does not by itself ensure explicit reporting of the encoded state. Note that its action expert conditions on the full VLM context, so probes and verbal outputs remain indirect measures of the information used for control.

Across the evaluated adaptation settings, probe and verbal performance can both improve, but not necessarily to the same extent. This is particularly clear on T1, where for Qwen3-VL, adaptation increases both probe (41.1$\rightarrow$69.9\%) and verbal performance (30.5$\rightarrow$71.0\%), whereas for Alpamayo-1.5 the probe changes only modestly (60.7$\rightarrow$63.6\%) while verbal performance improves substantially (29.7$\rightarrow$52.9\%). Thus, task-specific adaptation can strengthen the linearly decodable representation, but in some settings the larger effect is on verbal readout, making it useful to finetune a driving foundation model on tasks.

The T2 adapter-placement ablation further shows that substantial gains can arise from the language pathway alone: LLM-only adaptation nearly matches full adaptation for Qwen3-VL and yields the strongest verbal result for Alpamayo-1.5, while vision-encoder-only adaptation is weaker for both models in this setup. Together with the layerwise probes in Figure~\ref{fig:layerwise-probes}, this underscores that substantial speed-readout processing can occur within the LLM.

We additionally adapt Qwen3-VL directly to continuous speed estimation to test whether the readout improvements observed for categorical prediction also extend to metric supervision (Fig.~\ref{fig:kmh_scatter_ft}). Adaptation improves both verbal and probe estimates across all three tasks. For T2, verbal MAE decreases from 10.4 to 2.9 km/h and $\rho$ increases from 0.53 to 0.95, closely matching the probe (3.0 km/h, $\rho{=}0.94$). T3 likewise improves from 10.6 to 6.7 km/h verbally and from 7.0 to 5.2 km/h for the probe. T1 improves more modestly, from 15.5 to 8.8 km/h verbally and from 9.1 to 8.1 km/h for the probe, indicating that its remaining error is not attributable to readout alone.

Notably, the adapted verbal predictions remain strongly discretized despite continuous supervision: the model emits only 44 distinct integer speeds on T2, with 82\% of nonzero predictions taking even values, while several speeds that occur frequently in the training targets are never produced. In contrast, a continuous regression probe on the same adapted hidden states recovers a smooth range of speeds. Thus, the hidden representation supports finer-grained continuous speed decoding than is expressed through the verbal interface.

Cross-format transfer is asymmetric. Continuous-km/h adaptation largely preserves three-class performance (T2: 88.7\% vs. 90.1\% matched, T3: 67.9\% vs. 72.4\% matched macro-F1), whereas categorical adaptation transfers poorly to metric output (15.5/22.6 km/h MAE for T2/T3), often reverting to a small set of canonical values. Transfer is therefore substantially stronger from the finer metric formulation to the coarser categorical one than in the reverse direction.

Table~\ref{tab:shuffle_ft} compares ordered and shuffled five-frame inputs before and after adaptation. For T2, the ordered--shuffled gap is small in the zero-shot setting (0.8 pp for both verbal output and probe) and larger after adaptation (3.1 pp verbal, 5.4 pp probe), suggesting some increased dependence on frame order under adaptation. For T3, adaptation raises overall performance for both verbal output and probes, while reducing the ordered--shuffled gap: verbal performance remains nearly unchanged by shuffling (72.4\% vs. 72.2\%), and the probe gap narrows from 5.0 to 1.8 pp. Thus, most of the future-speed gain persists when chronology is disrupted and cannot be attributed to improved temporal grounding.

Overall, the fixed QLoRA recipe demonstrates substantial task-specific adaptability, particularly for ego-speed readout. However, continuous T1 estimation remains comparatively weak and temporal-order sensitivity limited. The results therefore demonstrate substantial correctability under direct supervision, but do not establish a general multi-agent temporal representation.

\section{DISCUSSION}
\label{sec:discussion}

\begin{table*}[t]
\centering
\caption{Summary of the main findings, supporting evidence, and potential mitigations. Related findings indicate results that partially improve or qualify the respective behavior.}
\label{tab:discussion_overview}
\scriptsize
\setlength{\tabcolsep}{5pt}
\renewcommand{\arraystretch}{1.05}

\begin{tabularx}{\textwidth}{
    @{}
    >{\raggedright\arraybackslash}p{0.39\textwidth}
    !{\color{gray!35}\vrule width 0.4pt}
    >{\raggedright\arraybackslash}p{0.15\textwidth}
    !{\color{gray!35}\vrule width 0.4pt}
    >{\raggedright\arraybackslash}X
    @{}
}
\toprule
\textbf{Main finding} &
\textbf{Evidence} &
\textbf{Potential mitigation / related finding} \\
\midrule

\textbf{Weak agent-specific encoding (T1).}
Surrounding-agent speed is not reliably encoded in an agent-specific representation by the VLM foundation models.
&
\subsecref{sec:representations}{sec:probe_main}
(Table~\ref{tab:probe_main});
\subsecref{sec:representations}{sec:probe_crosstask}
(Fig.~\ref{fig:probe_crosstask});
Appendix~\ref{app:t1-controls}
(Tables~\ref{tab:t1-box-control} and~\ref{tab:t1-type-control});
Section~\ref{sec:lora}
(Table~\ref{tab:ft-results})
&
\textit{Agent-centric temporal or relative-motion supervision.}
Alpamayo-1.5's stronger zero-shot T1 probe and the substantial probe gain of adapted Qwen3-VL suggest that driving- and task-specific supervision can strengthen agent-speed representations
(Section~\ref{sec:lora}, Table~\ref{tab:ft-results}), while continuous T1 estimation on our non-optimized training remains weak (Fig.~\ref{fig:kmh_scatter_ft}).
\\
\rowrule

\textbf{Representation--verbalization gap (T2).}
Current ego speed is strongly linearly accessible in hidden states but only incompletely reflected in verbal outputs.
&
\subsecref{sec:results}{sec:results_verbal}
(Table~\ref{tab:main_results});
\subsecref{sec:representations}{sec:probe_main}
(Table~\ref{tab:probe_main});
\subsecref{sec:representations}{sec:kmh-scatter}
(Fig.~\ref{fig:kmh-scatter});
Section~\ref{sec:lora}
(Table~\ref{tab:ft-results})
&
\textit{Improve readout while explicitly supervising temporal motion cues.}
Task-specific adaptation largely closes the readout gap and increases
temporal-order sensitivity
(Section~\ref{sec:lora}, Table~\ref{tab:ft-results};
Fig.~\ref{fig:kmh_scatter_ft};
Table~\ref{tab:shuffle_ft}).
\\
\rowrule

\textbf{Planning performance can mask weak perceptual grounding (T3).}
T3 predictions and representations can achieve nontrivial target agreement despite limited sensitivity to chronological motion evidence, yet improve strongly when accurate current ego speed is supplied. Thus, planning scores alone can understate the importance of reliable dynamic-state perception.
&
\subsecref{sec:results}{sec:results_verbal}
(Table~\ref{tab:main_results});
\subsecref{sec:results}{sec:ego-speed-hint}
(Fig.~\ref{fig:false-hint})
\subsecref{sec:results}{sec:frame_order}
(Fig.~\ref{fig:shuffle});
&
\textit{Evaluate planning jointly with its perceptual prerequisites.} Improving explicit recovery and verification of the current dynamic state may strengthen planning-related predictions, while target accuracy alone should not be interpreted as evidence of perceptual or temporal grounding.
\\
\rowrule

\textbf{Limited use of chronological history.}
Ordered and shuffled five-frame inputs perform similarly in most zero-shot settings, and rare gains from additional history often survive shuffling.
&
\subsecref{sec:results}{sec:frame_order}
(Fig.~\ref{fig:shuffle})
&
\textit{Explicit temporal supervision.}
Task-specific adaptation produces limited evidence of increased order sensitivity (Section~\ref{sec:lora}, Table~\ref{tab:shuffle_ft}), leaving temporal-order or motion supervision as an open direction.
\\
\rowrule

\textbf{Prompt-supplied ego state is integrated zero-shot and dominates visual cues.}
Without task-specific training on ego-state inputs, VLMs already use numerical ego-speed hints to improve T2/T3. Counterfactual hints strongly mislead both tasks, showing that prompt-supplied state is readily incorporated but weakly cross-checked against visual evidence.
&
\subsecref{sec:results}{sec:results_verbal}
(Table~\ref{tab:main_results});
\subsecref{sec:results}{sec:ego-speed-hint}
(Fig.~\ref{fig:false-hint})
&
\textit{Conflict-aware multimodal fusion and uncertainty calibration.} Inconsistencies between prompt-supplied state and visually inferred state should be checked and biased prompts should be rejected when inconsistent. Counterfactual hints provide a direct robustness test for such failures.
\\
\rowrule

\textbf{Formulation-sensitive and biased verbal outputs.}
Paired categorical and metric predictions can contradict, and several models exhibit persistent class preferences or output collapse. Thus, different output interfaces do not reliably expose a consistent speed estimate.

&
\subsecref{sec:results}{sec:output_formulation}
(paired cross-formulation analysis;
Tables~\ref{tab:main_results}, \ref{tab:binary_f1}, and~\ref{tab:kmh_mae};
Fig.~\ref{fig:violin-class-dist});
Section~\ref{sec:lora} 
&
\textit{Cross-format supervision with shared state representations.}
Task-specific adaptation improves the supervised format but transfers poorly across formulations. Cross-format training and shared state representations can be directions for further study.
\\
\rowrule

\textbf{Driving specialization does not eliminate interface-sensitive verbalization.}
Alpamayo-1.5 encodes surrounding-agent speed more strongly than Qwen3-VL, but this advantage is task-specific and not consistently reflected in verbal outputs. This suggests that stronger internal motion representations do not ensure reliable readout.
&
Section~\ref{sec:lora}
(Table~\ref{tab:ft-results})
&
\textit{Representation--readout alignment for driving-specialized models.}
Driving-specific representations should support explicit state readouts, with latent-state quality and verbal-output reliability evaluated separately.
\\

\bottomrule
\end{tabularx}
\end{table*}

Table~\ref{tab:discussion_overview} summarizes the main failure modes identified across the behavioral, probing, and adaptation analyses, together with potential mitigation strategies and the experimental findings that partially improve the observed behavior.

The perceptual tasks reveal distinct failure modes. Surrounding-agent speed (T1) is only weakly encoded in an agent-specific form, whereas current ego speed (T2) is substantially more accessible in hidden representations than in verbal outputs. T3 likewise exhibits a representation--verbalization gap, although its interpretation is less direct because the recorded trajectory represents one of several plausible short-horizon outcomes, whereas T1 and T2 have a single ground truth. Task-level performance alone cannot distinguish failures in state encoding, temporal integration, and verbal readout.

Temporal perturbations further show that the available motion evidence is represented, but not used reliably. T2 changes little when frame order is disrupted despite depending on ego-induced scene motion, and T3 achieves nontrivial target agreement with similarly limited sensitivity to chronology. Supplying the correct current ego speed improves T2 for every model and T3 for nearly all models. This shows that current-state information is essential, even when it is not reliably recovered from the visual sequence, and highlights the importance of assessing VLMs' underlying physical-state representations rather than focusing solely on finetuning for the final planning objective.
The counterfactual hints expose a robustness risk, showing that incorrect ego-speed information collapses T2 performance and substantially degrades T3. The models readily incorporate prompted state without rejecting erroneous information based on the visual evidence, creating vulnerability to upstream state errors or adversarially corrupted inputs.

Finetuning improves readout and representations, with modest frame-order sensitivity for current ego speed, while future-speed gains persist under shuffling, indicating improved target alignment without stronger temporal integration.
Overall, the VLMs are sensitive to prompting and output format, missing a prompt- and output-format-independent speed representation that is physically consistent. They exhibit task- and model-specific biases, making it difficult to interpret the failure modes.

This evaluation is not intended as a benchmark ranking, and no model consistently dominates across tasks or formulations. For exploratory analyses, multiple models provide broadly competitive baselines here, including Cosmos-R1/R2, Qwen3-VL, and InternVL2.5. The toolkit instead helps expose model-specific strengths and failure modes, such as LLaVA-Video's collapse in several categorical settings despite more usable continuous outputs. Because pretraining data are not fully known, such patterns may not generalize across datasets, but their variation shows that no single task or output formulation is sufficient to assess temporal perception for autonomous driving.

Our findings have direct implications for VLM-based planning. Plausible future-speed outputs may coexist with insufficient perception of the current dynamic state, while accurate state information, when supplied externally, can substantially improve those outputs. Open-loop accuracy or VQA performance should therefore not be treated as sufficient evidence of visually grounded planning. 
Evaluation should test whether models recover relevant state variables from vision, use ordered temporal evidence, and remain robust to corrupted auxiliary inputs. More broadly, dynamic competence should be assessed across four stages: state encoding, whether relevant physical quantities are represented internally and consistently; temporal integration, whether they are derived from ordered motion evidence; readout, whether internal state is faithfully exposed in the output; and action use, whether that state appropriately influences predictions. Distinguishing these stages is important because downstream success does not guarantee reliable perception or grounding.

This perspective extends beyond autonomous driving to embodied and robotic systems, where agents likewise need to recover dynamic state and use it for action. Evaluations of embodied VLMs and VLA's should complement task-level success with tests of state recovery, temporal evidence use, and robustness to inconsistent auxiliary information.

\subsection{LIMITATIONS}

The study uses five front-camera frames at 2Hz from nuScenes. Longer histories, higher frame rates, wider views, or multi-camera input may provide stronger motion evidence, and evaluation on additional datasets is needed to establish generality. The model set mainly covers open-weight VLMs of approximately 4--10B parameters as representatives of common models used for driving research. We note that larger or proprietary systems may behave differently. Adaptation is evaluated on few models with one non-optimized QLoRA configuration and should be interpreted diagnostically.
Furthermore, linear probes measure decodability rather than causal use. The binary, categorical, and numerical tasks also change the output formulation rather than testing prompt paraphrases. Finally, T3 uses a single recorded future in an open-loop setting and therefore measures agreement with one trajectory rather than the full quality of a driving proposal.

\section{CONCLUSION}
\label{sec:conclusion}
We investigated velocity understanding as a controlled diagnostic of dynamic perception in driving VLMs. Across the evaluated general-purpose and Physical-AI models, surrounding-agent speed is weakly encoded in an agent-specific form, while current ego speed is often internally decodable but poorly reported. Accurate ego-speed information substantially improves both current- and future-speed outputs, yet counterfactual hints are not reliably rejected using the video. Our task-specific adaptation improves reporting,but does not consistently strengthen temporal-order sensitivity. Our findings show that plausible or accurate planning-related outputs do not by themselves establish reliable visual-temporal grounding. Evaluating driving VLMs therefore requires separate tests of state recovery, temporal integration, output readout, and robustness to inconsistent evidence.

\appendices
\section{T1 CONTROL EXPERIMENTS}
\label{app:t1-controls}

The following controls test whether weak surrounding-agent speed estimation mainly reflects difficulty associating the target across frames or recognizing the marked vehicle.

In the standard T1 condition, the target vehicle is marked in the
current frame. We additionally mark the same target in all five frames to test whether persistent target localization improves speed estimation. Table~\ref{tab:t1-box-control} shows that marking the target vehicle in all five frames does not improve T1 speed estimation overall. Performance changes are small and inconsistent across models, with a mean macro-F1 change of $-0.6$ points relative to marking the target only in the current frame. This suggests that difficulty maintaining the target association across frames is unlikely to be the primary cause of the weak T1 speed-estimation performance.

\begin{table}[!t]
\centering
\caption{T1 speed macro-F1 (\%) when the target is marked only in the current frame or in all five frames. Marking all frames provides no consistent improvement. $^\dagger$MiniCPM-V produces unparsable outputs for 0.8\% for current marked frame and 1.4\% for all marked frames. $\Delta$ denotes all-frame minus current-frame macro-F1.}
\label{tab:t1-box-control}
\setlength{\tabcolsep}{3.5pt}
\renewcommand{\arraystretch}{1.0}
\footnotesize
\begin{tabular}{@{}lrrr@{}}
\toprule
\textbf{Model} &
\textbf{5F current} &
\textbf{5F all} &
$\boldsymbol{\Delta}$ \\
\midrule
Cosmos-Reason2   & 21.8 & 22.9 & +1.1 \\
Cosmos-Reason1   & 42.2 & 43.2 & +1.0 \\
Qwen2.5-VL       & 39.1 & 38.5 & -0.6 \\
Qwen3-VL         & 30.5 & 26.1 & -4.4 \\
LLaVA-Video       & 14.2 & 13.6 & +0.6 \\
InternVL2.5       & 41.4 & 41.4 & +0.0 \\
InternVL2         & 39.4 & 41.0 & +1.6 \\
Phi-3.5-Vision    & 26.2 & 25.9 & -0.3 \\
MiniCPM-V$^\dagger$ & 35.2 & 31.7 & -3.5 \\
\midrule
Mean       & 32.2 & 31.6 & -0.6 \\
\bottomrule
\end{tabular}
\end{table}

\subsection{VEHICLE-TYPE RECOGNITION}
\label{app:t1-type-control}

\begin{table}[!htbp]
\centering
\caption{Vehicle-type recognition for the marked T1 target under five-frame input. Type recognition is substantially stronger than T1 speed estimation, indicating that weak agent-speed performance cannot be explained by target recognition alone. Unparsable type responses are scored as errors. Cosmos-Reason1 (9.1\%), MiniCPM-V (3.5\%), and Qwen2.5-VL (3.1\%) had non-negligible unparsable rates; all other models $\leq$0.1\%.}
\label{tab:t1-type-control}
\setlength{\tabcolsep}{4pt}
\renewcommand{\arraystretch}{1.0}
\footnotesize
\begin{tabular}{@{}lrr@{}}
\toprule
\textbf{Model} &
\textbf{Accuracy} &
\textbf{Macro-F1} \\
\midrule
Cosmos-Reason2    & 76.2 & 34.0 \\
Cosmos-Reason1    & 69.6 & 48.0 \\
Qwen2.5-VL        & 73.6 & 49.3 \\
Qwen3-VL          & 78.5 & 43.8 \\
LLaVA-Video        & 75.8 & 36.7 \\
InternVL2.5        & 70.9 & 44.8 \\
InternVL2          & 75.8 & 37.8 \\
Phi-3.5-Vision     & 74.6 & 34.5 \\
MiniCPM-V          & 72.6 & 37.9 \\
\bottomrule
\end{tabular}
\end{table}

T1 jointly requests vehicle type and speed class. We therefore evaluate vehicle-type recognition separately in Table~\ref{tab:t1-type-control} to test whether poor speed estimation mainly reflects failure to identify the marked target.
The gap between accuracy and macro-F1 reflects the strong imbalance in vehicle types and the models' preference for common categories such as \emph{car}. Overall, vehicle-type macro-F1 is higher than T1 speed macro-F1 for eight of the nine evaluated  models, with mean scores of 40.8 and 32.2, respectively. InternVL2 is the only exception, achieving slightly lower macro F1 for vehicle type than for speed (37.8 vs.\ 39.4). These results suggest that weak T1 speed estimation cannot generally be attributed to vehicle-type recognition errors alone and point to additional difficulty in estimating agent-specific dynamic state.

\section{COMPLEMENTARY NUMERICAL SPEED-ESTIMATION RESULTS}
\label{app:metric-results}

Table~\ref{tab:kmh_rmse} complements the MAE results reported in
Table~\ref{tab:kmh_mae} with root mean squared error (RMSE). While MAE weights all absolute deviations equally, RMSE places greater emphasis on large prediction errors and therefore provides an additional view of the error distribution. Consistent with the MAE results, additional frames do not yield a systematic improvement across models and tasks, and performance remains strongly model- and task-dependent. Differences in the model ranking indicate that some models incur fewer large errors despite similar average absolute errors.
Most notably, for T2 with 5F, Cosmos-R1 achieves the lowest MAE (10.2 km/h) (see Table~\ref{tab:kmh_mae}), whereas Qwen3-VL achieves the lowest RMSE (13.7 km/h compared with 15.5 km/h for Cosmos-R1). Thus, for T2 with 5F, Cosmos-R1 is slightly more accurate on average, but Qwen3-VL exhibits fewer extreme prediction errors, resulting in a lower RMSE. Similarly, for T3 with 1F, Cosmos-R1 has a slightly lower MAE than MiniCPM-V (13.4 vs. 13.8 km/h), but a higher RMSE (18.1 vs. 16.9 km/h).

\begin{table}[!t]
\centering
\caption{Root mean squared error (RMSE) in km/h for numerical speed estimation on T1--T3 under single-frame (1F) and five-frame (5F) input. Lower is better. LLaVA-1.5 receives only the current frame. Parsing rate is equivalent to MAE results: 99.8\% for Phi-3.5-V on T1--5F and for MiniCPM-V$^\dagger$ on T2--1F (98.5\%), T2--5F (77.5\%), and T3--1F (82.6\%).}
\label{tab:kmh_rmse}
\setlength{\tabcolsep}{3.2pt}
\renewcommand{\arraystretch}{0.98}
\footnotesize
\begin{tabular}{@{}lcccccc@{}}
\toprule
&
\multicolumn{2}{c}{\textbf{T1 surrounding}} &
\multicolumn{2}{c}{\textbf{T2 ego current}} &
\multicolumn{2}{c}{\textbf{T3 ego future}} \\
\cmidrule(lr){2-3}
\cmidrule(lr){4-5}
\cmidrule(l){6-7}
\textbf{Model}
& \textbf{1F} & \textbf{5F}
& \textbf{1F} & \textbf{5F}
& \textbf{1F} & \textbf{5F} \\
\midrule
Cosmos-R2 & \textbf{17.4}  &  \textbf{14.5}  &  22.7  &  18.1  &  19.4  &  15.7  \\
Cosmos-R1 &  23.9  &  19.2  &  \textbf{18.7}  &  15.5  &  18.1  &  26.5  \\
Qwen2.5-VL &  18.2  &  21.1  &  22.0  &  15.9  &  16.6  &  17.9  \\
Qwen3-VL &  22.3  &  19.8  &  22.7  &  \textbf{13.7}  &  19.5  &  \textbf{14.0}  \\
LLaVA-Video &  18.7  &  18.6  &  22.7  &  22.3  &  19.2  &  18.7  \\
\midrule
InternVL2.5 &  17.5  &  17.3  &  19.6  &  20.6  &  16.0  &  17.1  \\
InternVL2 &  18.4  &  19.3  &  19.8  &  17.7  &  21.8  &  16.3  \\
Phi-3.5-V &  19.4  &  19.5  &  22.7  &  22.6  &  23.1  &  25.0  \\
MiniCPM-V$^\dagger$  &  19.4  &  18.9  &  22.3  &  20.5  & 16.9 &  \textbf{14.0}  \\
LLaVA-1.5 &  18.0  &  --  &  21.5  &  --  &  \textbf{15.2}  &  -- \\
\bottomrule
\end{tabular}
\end{table}

\begin{table}[!htbp]
\centering
\caption{Median Spearman's $\rho$ across the ten models (excluding LLaVA-1.5 from the 5F+S condition) between T1 numerical predictions and the target-vehicle or ego speed. Supplying ego speed shifts predictions strongly toward ego state rather than the queried vehicle.}
\label{tab:t1-ego-anchoring}
\small
\begin{tabular}{@{}lcc@{}}
\toprule
Condition & Target speed & Ego speed \\
\midrule
1F   & 0.27 & 0.29 \\
1F+S & 0.32 & 0.77 \\
5F+S & 0.34 & 0.84 \\
\bottomrule
\end{tabular}
\end{table}

\section{EGO-SPEED ANCHORING}
\label{app:t1-ego-anchoring}

To test whether the ego-speed hint is integrated with agent-specific motion information or directly influences the surrounding-agent prediction, we compare T1 numerical outputs with both the target-vehicle speed and the ego speed. Table~\ref{tab:t1-ego-anchoring} reports the median Spearman correlation across the ten evaluated models. Without a hint, correlations with target and ego speed are similar. With supplied ego speed, T1 predictions correlate substantially more strongly with ego speed than with the queried vehicle's speed, indicating anchoring on the supplied ego state.

\section*{ACKNOWLEDGMENT}
The research leading to these results is funded by the German
Federal Ministry for Economic Affairs and Energy (BMWE) within the project “NXT GEN AI METHODS -- Generative Methoden für Perzeption, Prädiktion und Planung” (grant no. 19A23914M) and 
the German Federal Ministry of Research, Technology and Space (BMFTR) within the project ADRIVE-GPT (grant no. 13FH544KA2). 
The authors gratefully acknowledge the scientific support and HPC resources provided by the Erlangen National High Performance Computing Center (NHR@FAU) of the Friedrich-Alexander-Universität Erlangen-Nürnberg (FAU) under the BayernKI project v103fe. BayernKI funding is provided by Bavarian state authorities.
The authors are solely responsible for the content of this publication.

Generative AI tools were used in this work for orchestrating and monitoring cluster
runs of experiments, statistical analysis, and reviewing the writing style and grammar of the paper. They were not used to generate synthetic data, and no
experimental measurement reported here was produced by a language model. All
numbers come from executed training and evaluation runs whose logs are retained.
We have reviewed all AI-assisted work. Every result in this paper was regenerated
from raw evaluation logs. Each experimental arm asserts its own initialization
programmatically before scoring and the primary analysis (metric, regime and
sample size) was fixed in advance of the corresponding runs. We take full responsibility for the final content of this
work, including text, claims, and artifacts produced with the aid of generative
AI.

\bibliographystyle{ieeetr}
\bibliography{references.bib}

\begin{IEEEbiography}[{\includegraphics[width=1in,height=1.25in,clip,keepaspectratio]{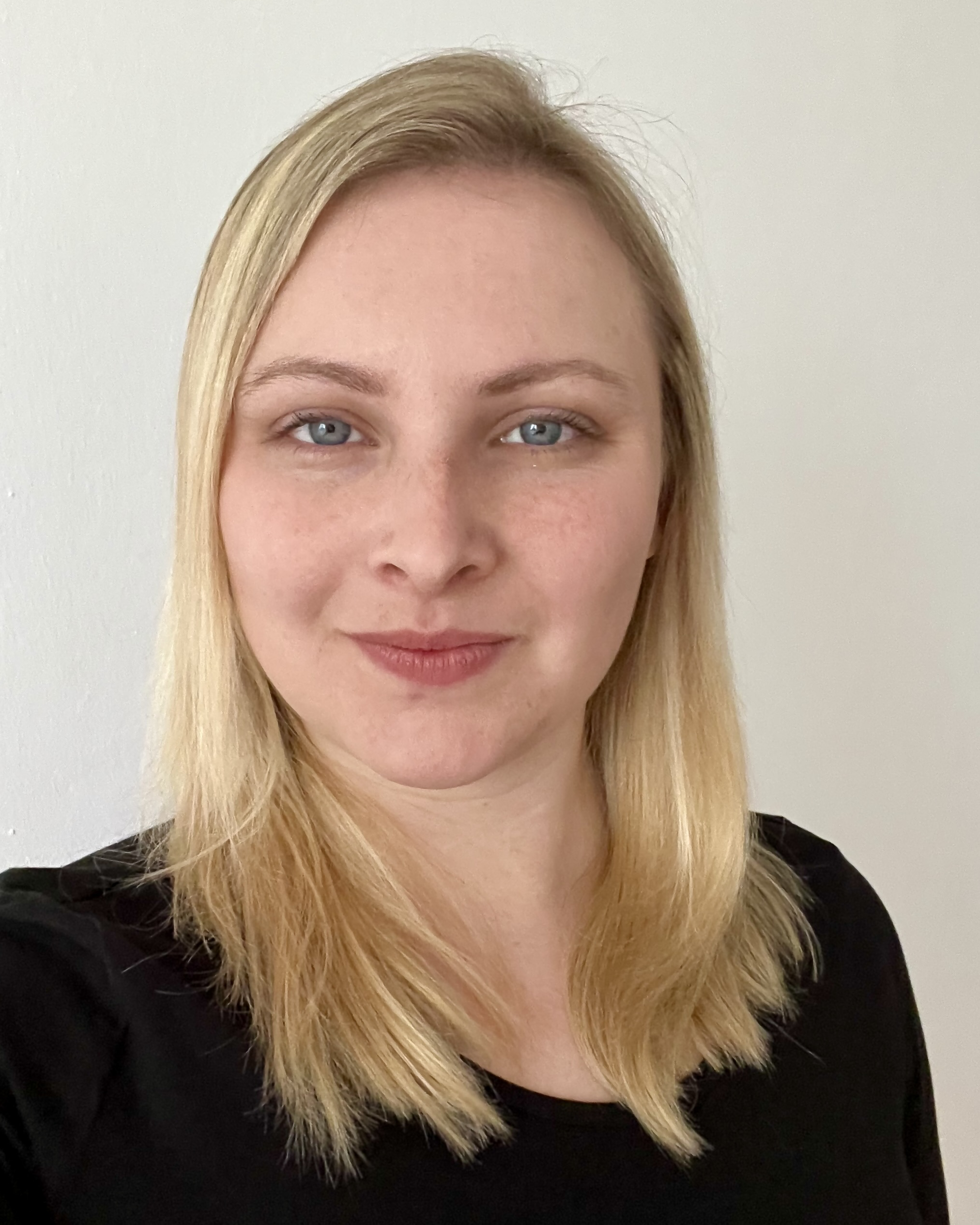}}]{Katharina Winter } received her M.Sc. in Media Informatics at LMU Munich in 2023.
She is currently pursuing her Ph.D. at Munich University of Applied Sciences in the Intelligent Vehicles Lab.  Her research focuses on multimodal large language and vision--language models for autonomous driving, with particular emphasis on end-to-end trajectory planning, spatio-temporal scene understanding, generative world models, and explainable AI.
\end{IEEEbiography}

\begin{IEEEbiography}[{\includegraphics[width=1in,height=1.25in,clip,keepaspectratio]{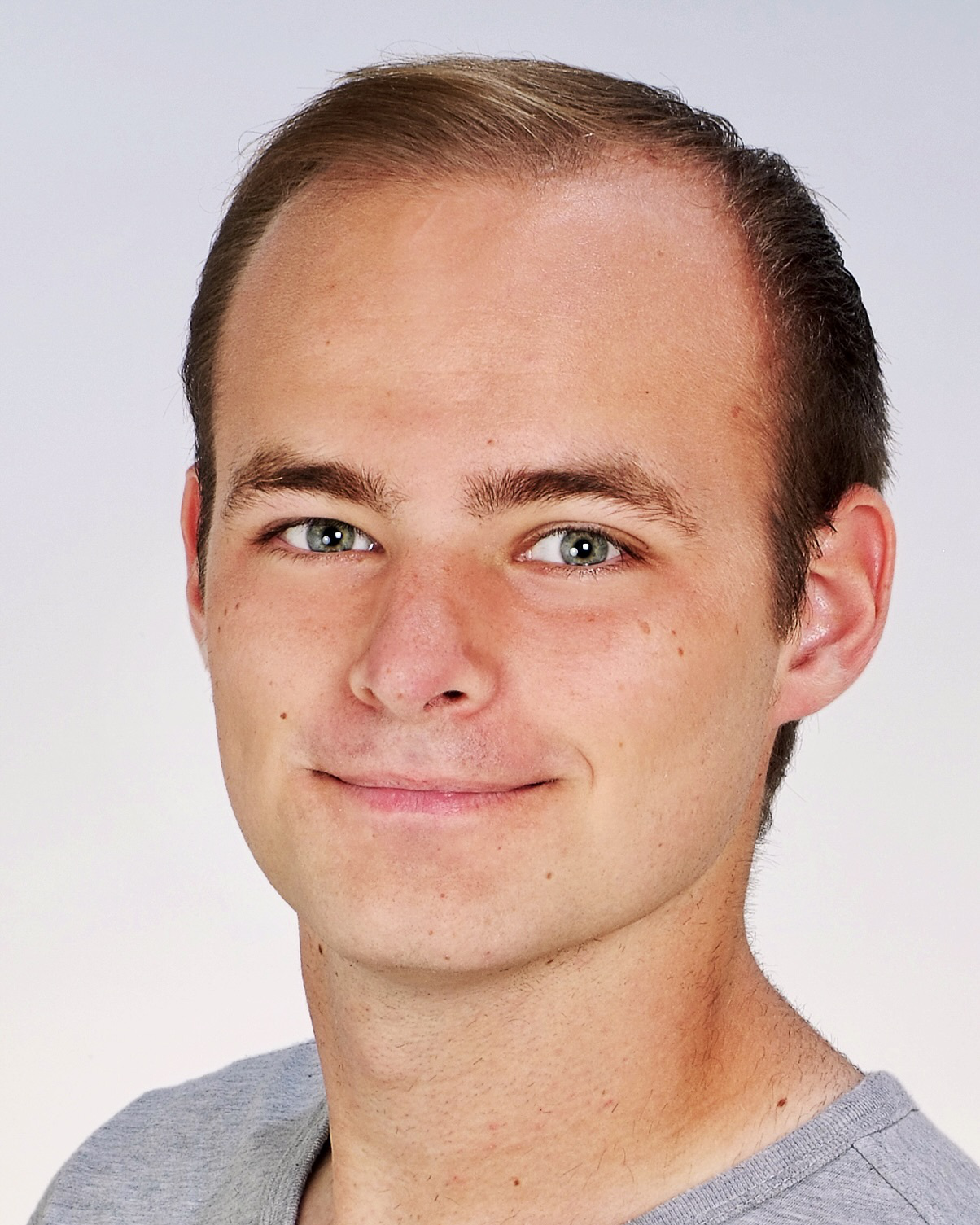}}]{Stefan Englmeier } received his B.Sc. degree in scientific computing and M.Sc. degree in computer science from Munich University of Applied Sciences, Munich, Germany in 2024. He is currently pursuing his Ph.D. at Munich University of Applied Sciences in the Intelligent Vehicles Lab, Munich, Germany.
His research focuses on autonomous driving, specifically on trajectory planning using large language and world models to improve decision-making and robustness in complex driving environments.
\end{IEEEbiography}

\begin{IEEEbiography}[{\includegraphics[width=1in,height=1.25in,clip,keepaspectratio]{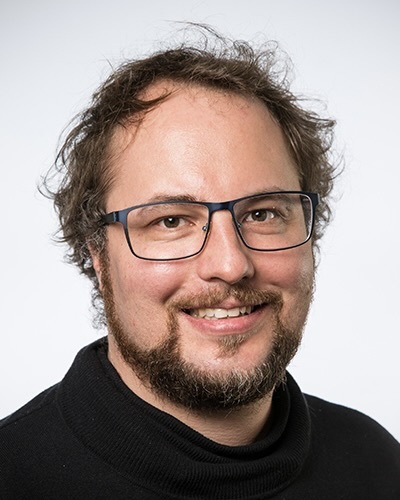}}]{Fabian
B. Flohr } received his Ph.D. degree in Computer Science from University of Amsterdam, The Netherlands, in 2018. From 2012 to 2022, he was with Mercedes-Benz Research and Development in Stuttgart, Germany, where he focused on automated driving. During this time, he served as Technical Manager for Vulnerable Road User (VRU) Protection, shaping and coordinating the strategic and technical direction of the topic across multiple international teams. Since 2022, he has been a Full Professor of Machine Learning at Munich University of Applied Sciences, where he leads the Intelligent Vehicles Lab.
\end{IEEEbiography}

\vfill

\end{document}